\documentclass{article}

\usepackage[final]{neurips_2026}
\makeatletter
\providecommand{\@trackname}{}
\makeatother
\usepackage{booktabs}
\usepackage{multirow}
\usepackage{adjustbox}
\usepackage{array}
\usepackage[utf8]{inputenc}
\usepackage[T1]{fontenc}
\usepackage{microtype}
\usepackage{hyperref}
\usepackage{amsmath}
\usepackage{cleveref}
\usepackage{xurl} 
\usepackage{booktabs}
\usepackage{multirow}
\usepackage{amsfonts}
\usepackage{amssymb}
\usepackage{nicefrac}
\usepackage[table]{xcolor}
\usepackage{graphicx}
\usepackage{subcaption}
\usepackage{todonotes}
\usepackage[breakable]{tcolorbox}
\usepackage{fvextra}
\usepackage{wrapfig}
\definecolor{darkblue}{rgb}{0, 0, 0.5}
\definecolor{bestcolor}{rgb}{0.0, 0.45, 0.0}
\definecolor{ourpurple}{RGB}{230,220,245}
\definecolor{citecolor}{HTML}{0071BC}
\definecolor{linkcolor}{HTML}{ED1C24}
\tcbuselibrary{skins}
\newtcolorbox{promptbox}[1]{breakable,enhanced,colback=gray!5,colframe=gray!55,fonttitle=\bfseries,title=#1,left=4pt,right=4pt,top=4pt,bottom=4pt}
\hypersetup{colorlinks=true, citecolor=citecolor, linkcolor=linkcolor, urlcolor=darkblue}
\definecolor{secgray}{HTML}{ECECEC}
\definecolor{deltagreen}{HTML}{1B7E3C}
\newcommand{\dt}[1]{{\color{deltagreen}#1}}

\newcommand{\eg}{\emph{e.g}.}
\newcommand{\ie}{\emph{i.e}.}

\usepackage{subcaption}
\newcommand{\etal}{\emph{et al}.}
\usepackage{xspace}
\newcommand{\ours}{\textsc{ContextRL}}

\title{Context-Aware RL for Agentic and Multimodal LLMs}

\author{
  Peiyang Xu$^{1}$\thanks{Correspond to Peiyang Xu \texttt{<px4668@princeton.edu>}, Xingyu Fu \texttt{<xingyufu@princeton.edu>}.} \quad
  Bangzheng Li$^{2}$ \quad
  Sijia Liu$^{1}$ \quad
  Karthik R. Narasimhan$^{1}$ \\[1mm]
  {\bfseries
  Pramod Viswanath$^{1}$ \quad
  Prateek Mittal$^{1}$\thanks{Equal advising contribution.} \quad
  Xingyu Fu$^{1}$\footnotemark[\value{footnote}]} \\[2mm]
  $^{1}$\small\textit{Princeton University} \qquad $^{2}$\small\textit{UC Davis}
}

\begin{document}

\maketitle

\vspace{-10mm}
\begin{center}
\small
\href{https://xupy2003.github.io/ContextRL_Website/}{\raisebox{-0.15em}{\includegraphics[height=1em]{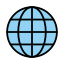}}\,\textbf{Website}} \quad
\href{https://github.com/xupy2003/ContextAwareRL}{\raisebox{-0.15em}{\includegraphics[height=1em]{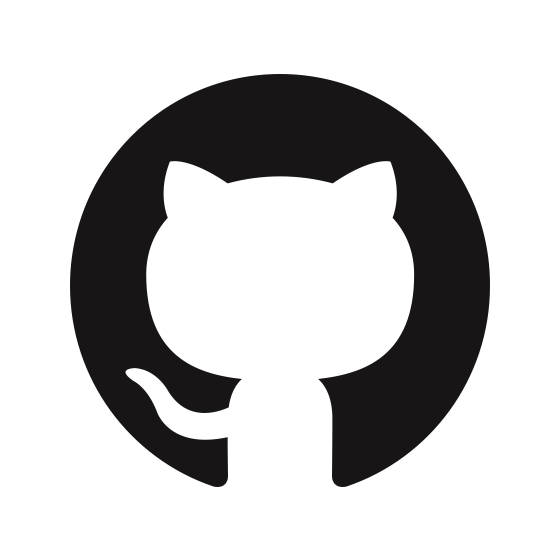}}\,\textbf{Code}} \quad
\href{https://huggingface.co/collections/xupy21/contextrl-models}{\raisebox{-0.15em}{\includegraphics[height=1em]{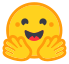}}\,\textbf{Model}} \quad
\href{https://huggingface.co/collections/xupy21/contextrl-datasets}{\raisebox{-0.15em}{\includegraphics[height=1em]{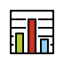}}\,\textbf{Dataset}}
\end{center}
\vspace{-2mm}

\begin{abstract}
    Large language models (LLMs) often fail when answering requires identifying a small but decisive piece of evidence within a long or complex context, such as a single line in a tool trace or a subtle detail in an image.
    We propose \ours, a context-aware reinforcement learning (RL) method that improves long-horizon reasoning and multimodal performance through an \emph{indirect} auxiliary objective.
    Instead of supervising only the final answer, \ours~presents the model with a query, an answer, and two highly similar contexts, and rewards it for selecting the context that supports the query--answer pair, thereby encouraging fine-grained grounding.
    We construct contrastive context data in two domains: for coding agents, trajectories serve as contexts, yielding 1k pairs built via condition filtering; for multimodal reasoning, images serve as contexts, yielding 7K pairs built via generative editing and similarity search.
    \ours~achieves average gains of +2.2\% over standard GRPO on 5 long-horizon benchmarks, and +1.8\% across 12 diverse visual question answering benchmarks. 
    To disentangle the effect of the proposed objective from that of additional data, we compare against data-augmentation baselines that repurpose the same contrastive contexts as standard query--context--answer examples. These baselines provide little to no improvement, showing that the gains arise from the proposed context-selection objective rather than from the contrastive data alone. Data and code will be publicly released.
\end{abstract}

%=====================================================================
\section{Introduction}
\label{sec:intro}
%=====================================================================

\begin{figure}[h]
\centering
\includegraphics[width=\linewidth]{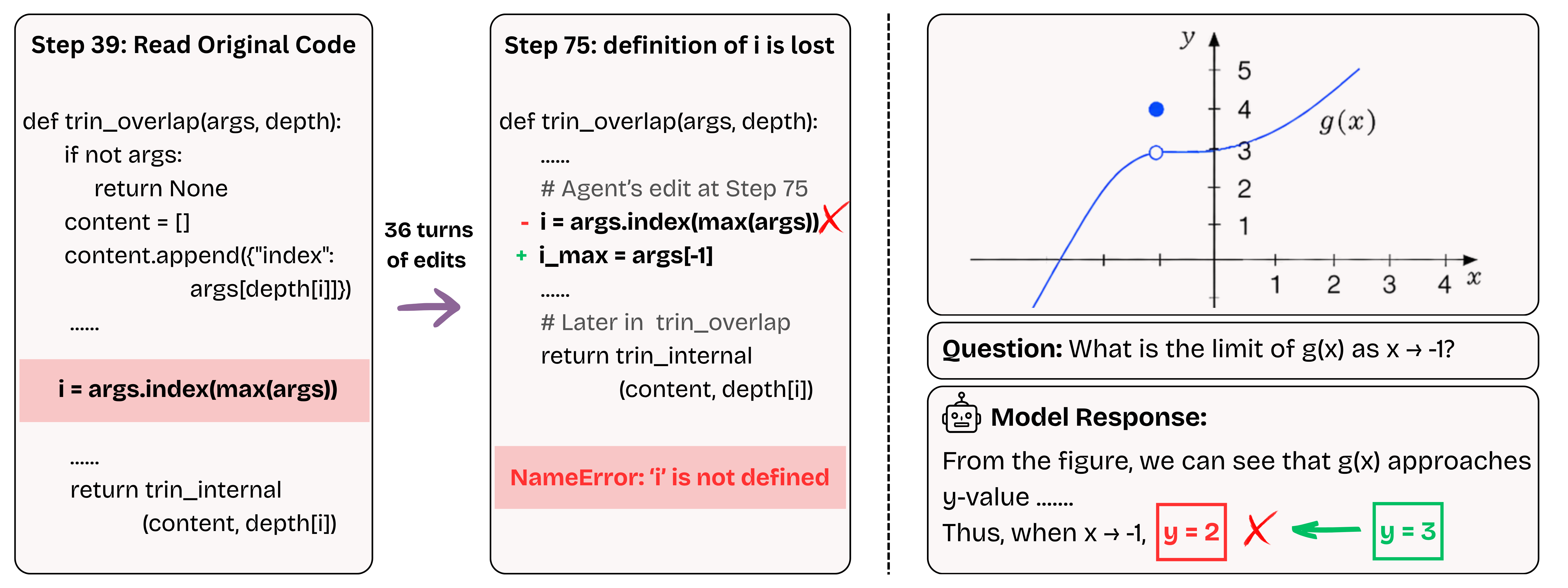}
\caption{\textbf{Context unawareness manifests across both agentic and multimodal systems.}
\textbf{Left:} In an agentic code-editing setting, the model has access to the relevant source file but fails to maintain consistency with the surrounding context across edits. As a result, it removes the definition of variable $i$ that is subsequently referenced, causing a runtime error.
\textbf{Right:} In a multimodal reasoning setting, the model fails to correctly ground its answer in the visual evidence. Although the relevant information is present in the figure, it misreads the $y$ value of $g(x)$ as $2$ rather than $3$ as $x \to -1$, leading to an incorrect prediction.}
\label{fig:unawareness_example}
\vspace{-3em}
\end{figure}

Large language models (LLMs) have progressed rapidly~\citep{team2023gemini, qwen3.5, singh2025openaigpt5card, grattafiori2024llama3herdmodels, deepseekai2026deepseekv4}, evolving from text completion systems into general-purpose reasoning engines~\citep{wei2023chainofthoughtpromptingelicitsreasoning, shao2024deepseekmath} capable of operating over rich, long-form contexts.
This progress has enabled a wide range of intelligent systems, including agentic systems~\citep{yao2023reactsynergizingreasoningacting,yang2024sweagent,anthropic2025claude47,2026openclaw,yao2023treethoughtsdeliberateproblem, shinn2023reflexionlanguageagentsverbal,wang2023voyageropenendedembodiedagent, schick2023toolformerlanguagemodelsteach} that interleave reasoning with tool use over extended horizons, and multimodal models~\citep{liu2023visualinstructiontuning, alayrac2022flamingovisuallanguagemodel,li2023blip2bootstrappinglanguageimagepretraining,team2024gemini,bai2025qwen3vltechnicalreport,singh2025openaigpt5card} that combine fine-grained perception with textual reasoning over high-dimensional inputs.

As these systems scale to increasingly complex tasks, their performance depends not only on reasoning ability, but also on grounding decisions in sparse yet decisive contextual evidence~\citep{liu2023lostmiddlelanguagemodels, mei2025surveycontextengineeringlarge}: an early observation in a long agent trajectory, a single line in a tool trace, or a subtle visual detail in a dense image. 
When such evidence is overlooked, models may make locally plausible but context-inconsistent decisions, such as modifying a pre-defined variable in code or missing a small visual cue needed to answer a question. 
A growing body of work~\citep{turpin2023languageUnfaithful, wang2026perceptionawarepolicyoptimizationmultimodal, chen2026longrlvrlongcontextreinforcementlearning, kamradt2023niah, bai2024longbench2} suggests that modern models often exhibit this failure mode. We refer to it as \textbf{\emph{context unawareness}}: the relevant information is available in the context, yet the model's prediction is not grounded in it. Qualitative examples are provided in \Cref{fig:unawareness_example}.

To quantify this failure mode, we introduce a controlled \emph{contrastive context probe}. We construct 200 contrastive context pairs from agentic trajectories and 200 from visual question answering (VQA) images, with examples shown in \Cref{fig:contrastive_examples}. Each example presents the model with 
\begin{wrapfigure}{r}{0.5\textwidth}
  \centering
  \includegraphics[width=0.48\textwidth]{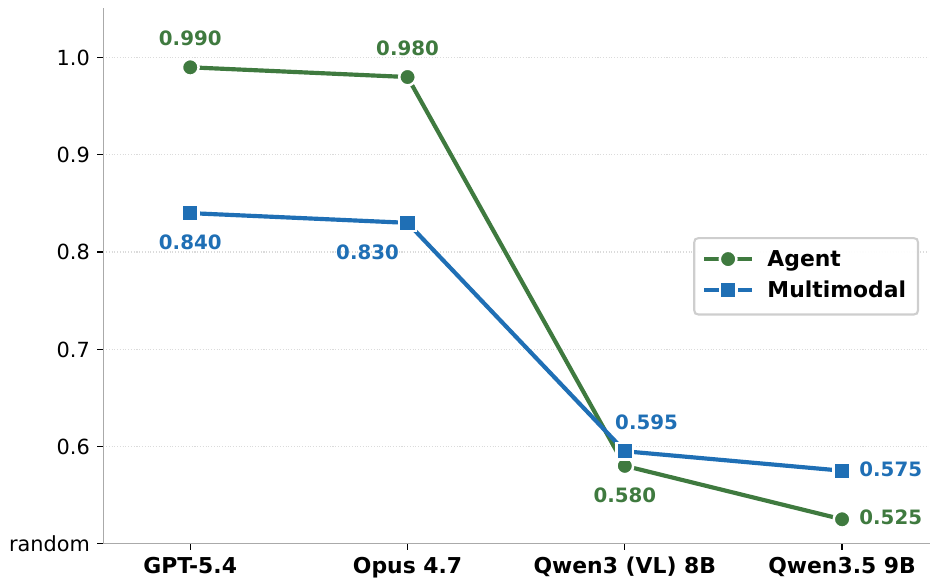}
  \caption{\small \textbf{Contrastive context selection given a query--answer pair.}
    GPT-5.4 and Claude Opus 4.7 perform reliably, while Qwen3 (VL) and Qwen3.5 remain close to random choice despite strong performance on standard benchmarks.}
  \label{fig:choice}
\vspace{-3em}
\end{wrapfigure}

a question, a candidate answer, and two closely matched contexts that support different answers; the model must select the context that justifies the candidate answer. As shown in \Cref{fig:choice}, this simple test reveals a roughly 40-point gap between proprietary models~\citep{singh2025openaigpt5card, anthropic2025claude47} and widely used open-source models~\citep{qwen3.5, yang2025qwen3technicalreport}. Notably, strong open-source models such as Qwen-3 (VL) 8B and Qwen-3.5 9B~\citep{qwen3.5, yang2025qwen3technicalreport} perform close to random choice, despite their competitive performance on standard benchmarks. 
These results suggest that strong benchmark performance can obscure failures in context grounding, where models struggle to identify the evidence that supports a given answer.

\begin{figure}[h]
\centering
\includegraphics[width=0.8\linewidth,keepaspectratio=false]{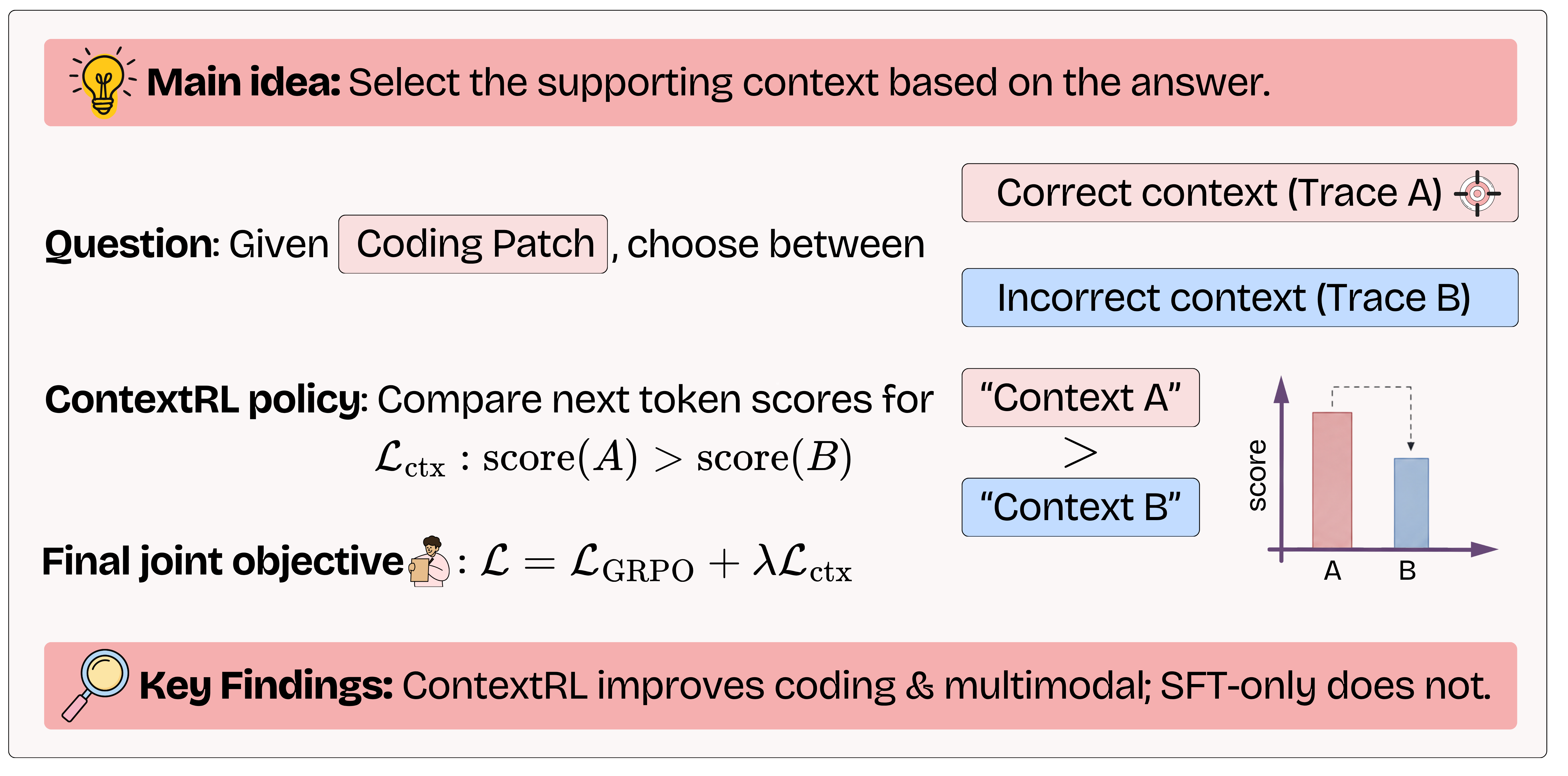}
\caption{
\textbf{Overview of \ours.} 
We augment GRPO with a context-awareness loss built from contrastive context pairs. For each query-answer pair $(Q,A)$, an anchor context $Context$ $A$ supports the answer, while a confounder $Context$ $B$ is superficially similar but inconsistent. Given the answer, the auxiliary objective trains the policy to select the supporting context over the confounder, improving context-aware reasoning in both coding and multimodal settings.}
\label{fig:pipeline}
\vspace{-1.5em}
\end{figure}

Motivated by this observation, we propose \textbf{Context-Aware Reinforcement Learning} (\ours), a post-training method that augments reinforcement learning (RL)  with explicit context-selection supervision. Whereas standard RL optimizes final answers through outcome rewards, \ours~adds an auxiliary objective that rewards the model for selecting the context that supports a given answer. 
As illustrated in \Cref{fig:pipeline}, we construct contrastive context pairs through condition search, generative editing, and similarity-based retrieval, and incorporate them into standard GRPO post-training~\citep{shao2024deepseekmath, yu2025dapoopensourcellmreinforcement} via a logit-level contrastive loss that favors the context aligned with the ground-truth answer.

Despite its simplicity, this auxiliary signal yields consistent improvements across 5 long-horizon and 12 multimodal benchmarks. 
For long-horizon experiments, we construct 1k contrastive trajectory pairs from SWE-smith trajectories~\citep{yang2025swesmithscalingdatasoftware} using condition filtering.
\ours~improves over the GRPO baseline by +3.2\% and +1.5\% on average across five agentic and long-context benchmarks, using Klear-AgentForge-8B~\citep{wang2025klearagentforgeforgingagenticintelligence} and Qwen3-8B~\citep{yang2025qwen3technicalreport} as base models, respectively. 
For multimodal experiments, we construct 7K contrastive image pairs from diverse sources through generative editing and similarity-based retrieval. Across 12 benchmarks, \ours~improves over standard GRPO~\citep{shao2024deepseekmath} by +2.0\% on Qwen2.5-VL-7B and +1.6\% on Qwen3-VL-8B, on average. 
To isolate the source of these gains, we compare against data-augmentation baselines that consume the same contrastive data through SFT or outcome-based RL. Strikingly, these baselines \emph{collapse or fail entirely}: supervised augmentation drives the long-horizon agent's resolve rate to as low as 0\%, while outcome-based augmentation adds essentially nothing. Only the proposed context-selection objective converts the same data into consistent gains, indicating that the improvement stems from the training objective rather than the contrastive data alone. 
These results show that a single context-selection objective improves both long-horizon reasoning and multimodal understanding, without architectural changes or large-scale human annotation.

Taken together, our findings suggest that failures in long-horizon reasoning and visual understanding often reflect a common limitation: models do not reliably ground their predictions in the evidence available to them. Our diagnostic probe exposes this limitation, and \ours~shows that it can be addressed with a lightweight context-selection objective. Comparisons with data-augmentation baselines further show that the key factor is not the contrastive data alone, but the way this signal is integrated into training. By decoupling \emph{what} answer to produce from \emph{which} context supports it, \ours~offers a simple, lightweight, and modality-agnostic mechanism for improving context grounding in complex, information-rich settings.

%=====================================================================
\section{\ours}
\label{sec:method}
%=====================================================================

We introduce \textbf{Context-Aware Reinforcement Learning} (\ours), a training framework that explicitly encourages models to ground their predictions in the provided context. Our key idea is to augment standard RL post-training with a \emph{context-selection} signal: instead of only rewarding correct outputs, we additionally train the model to identify \emph{which context supports a given answer}.
As illustrated in Figure~\ref{fig:pipeline}, we first construct \emph{contrastive context pairs} $(C^+, C^-)$ across both agentic and multimodal settings. Each pair is associated with a query $Q$ and an anchor answer $A$, where $C^+$ is the \emph{supporting} context for $A$ and $C^-$ is a minimally perturbed but \emph{confounding} alternative that instead supports a different answer. The context takes the form of an agent trajectory $\tau$ for agentic coding, or an image $I$ for multimodal tasks. We then jointly optimize the standard GRPO objective on the original task data and a context-awareness loss $\mathcal{L}_{\mathrm{CA}}$ on these contrastive pairs, which conditions on $(Q, A)$ and rewards the model for selecting $C^+$ over $C^-$. 

In the following, we describe the construction of contrastive data for agentic (\S\ref{sec:data_agentic}) and multimodal (\S\ref{sec:data_multimodal}) settings, followed by the joint training objective (\S\ref{sec:contrastive_rl}).

\subsection{Agentic Contrastive Context Pairs Construction}
\label{sec:data_agentic}

Figure~\ref{fig:contrastive_examples} (left) illustrates our pipeline for constructing contrastive trajectory pairs in the agentic setting. Each instance consists of a query $Q$ (a GitHub issue), an answer $A$ (the reference patch), and the trajectory $\tau$ capturing the agent’s reasoning trace, tool interactions, sandbox observations, and submitted patch. Correctly producing $A$ requires identifying which parts of the trajectory provide supporting evidence for the edit.

\vspace{1ex}\noindent\textbf{Mining contrastive trajectories.}
We construct contrastive trajectory pairs from the 66k trajectories~\citep{wang2025klearagentforgeforgingagenticintelligence} built from SWE-smith~\citep{yang2025swesmithscalingdatasoftware}. 
Our pipeline applies a cascade of increasingly restrictive filters: candidate trajectories must (i) originate from the same repository and commit, (ii) modify the same file, (iii) target the same function or class, and (iv) correspond to distinct but semantically related issues.
To prevent trivial leakage, patch contents inside edit commands are masked with \texttt{<PATCH\_MASKED>}.
These constraints yield pairs that are nearly identical at the token level: same repository, commit, file, and function, differing only in a small, decisive code region.Thus, selecting the correct trajectory requires understanding the context rather than exploiting surface statistics.
We then apply automatic verification with GPT~5.4~\cite{singh2025openaigpt5card} that explicitly screens for and rejects pairs that expose superficial shortcut cues or have ambiguous labels; cases the verifier marks \texttt{UNCERTAIN} are escalated to manual inspection. This cascade is deliberately aggressive: only \textbf{1k} high-quality contrastive trajectory pairs survive, \ie\ $1.5\%$ of the 66k source trajectories.
Details of the step-by-step procedure are provided in Appendix~\ref{app:agentic_mining} and the verification prompt is provided in Appendix~\ref{app:prompt_traj_verifier}.

\subsection{Multimodal Contrastive Context Pairs Construction}
\label{sec:data_multimodal}

\begin{figure}[t]
\centering
\includegraphics[width=\linewidth]{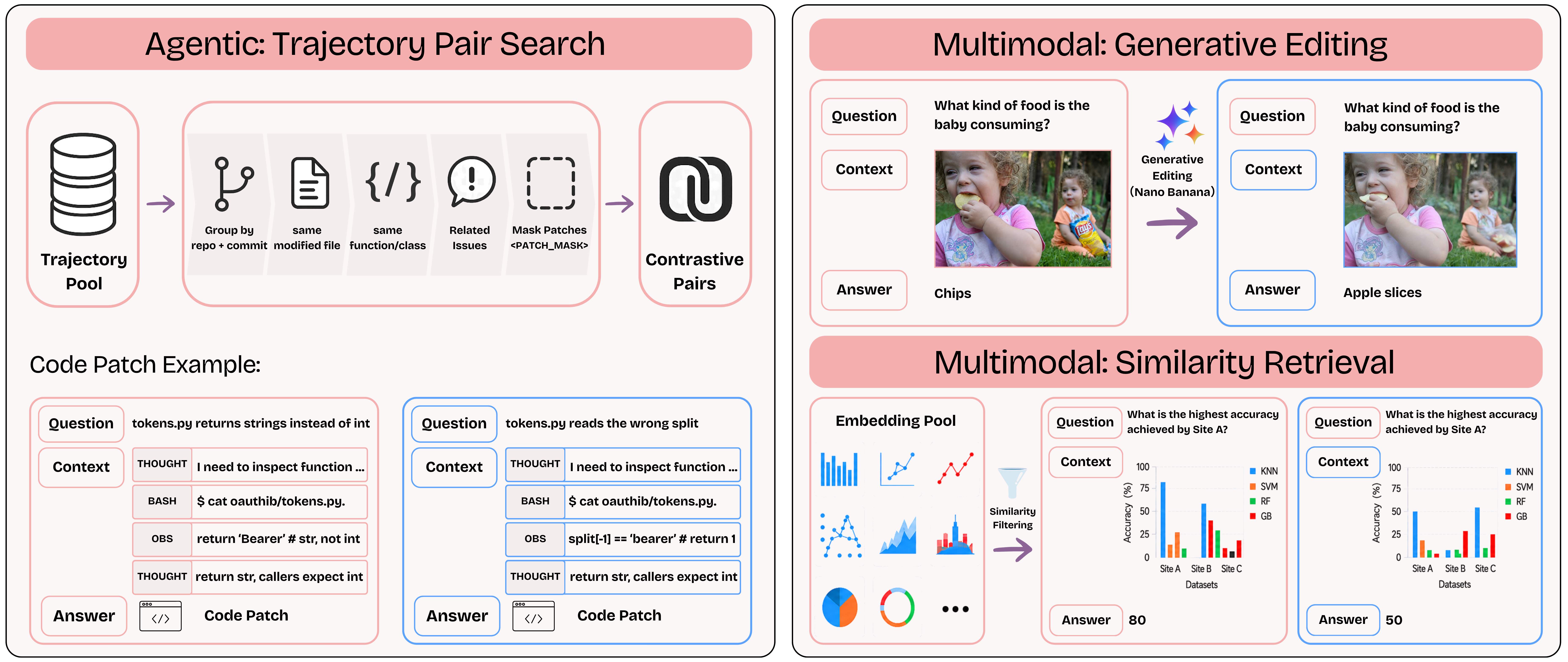}
\caption{\textbf{Contrastive context pairs construction pipeline.} \textbf{Left:} Step-by-step filtering to mine contrastive trajectory pairs for the \emph{agentic setting}. \textbf{Right:} Generative editing and similarity-based retrieval to mine contrastive image pairs for the \emph{multimodal setting}. A concrete $(C^{+}, C^{-})$ example is shown 
alongside each method. Details can be found in Section ~\ref{sec:data_agentic} and ~\ref{sec:data_multimodal}.}
\label{fig:contrastive_examples}
\end{figure}

Figure~\ref{fig:contrastive_examples} (right) shows our construction of contrastive image pairs through \emph{generative editing} and \emph{similarity-based retrieval}. We collect image, question, and answer triplets from five domains: charts, geometry, non-geometric math, science diagrams, and natural images, covering diverse patterns of visual grounding. Dataset sources are detailed in Appendix~\ref{app:multimodal_sources}. 

\vspace{1ex}\noindent\textbf{Generative editing for natural images.}
For natural images, we construct contrast pairs via controlled editing. Given $(I, Q, A)$, we generate a modified image $I'$ that preserves global scene structure while changing the answer to $A'$. This process consists of four stages:
(1) \emph{Instruction generation:} GPT~5.4 proposes an edit prompt that would change the answer;
(2) \emph{Image synthesis:} the edit is applied using Nano Banana 2~\citep{team2023gemini};
(3) \emph{Verification:} GPT~5.4 explicitly rejects edited images showing visible editing artifacts (blur, warping, broken object boundaries, implausible lighting, texture mismatch, or floating objects) or global restyling that could act as a whole-image ``tell'', and requires the edit to stay localized to the answer-relevant region while leaving question-irrelevant content unchanged (verification prompt in Appendix~\ref{app:prompt_edit_verifier}).
(4) Cases marked as \texttt{UNCERTAIN} by the verifier are additionally reviewed by the authors. This yields approximately 700 high-quality pairs from 2k candidates, a $\sim$65\% rejection rate.

\vspace{1ex}\noindent\textbf{Similarity-based retrieval for structured images.}
For structured or text-rich images, direct editing often violates underlying constraints (e.g., geometric consistency or chart semantics). Instead, we construct pairs by retrieval. We embed images using Qwen3-VL-Embedding 8B~\citep{qwen3vlembedding}, and for each $(I,Q,A)$ retrieve $(I',Q,A')$ such that $\cos(f_I(I), f_I(I')) \geq \alpha_I$ and $A \neq A'$, where we set $\alpha_I$ to 0.85 so that every retrieved negative is, by construction, highly visually similar to its positive yet supports a different answer. We then filter candidates using GPT~5.4 to remove semantically unrelated pairs or annotation artifacts (verification prompt in Appendix~\ref{app:prompt_edit_verifier}). This yields 6{,}300 high-quality pairs from over 200k candidates.

The final multimodal training set contains \textbf{7k} contrast image pairs, combining both strategies.

\vspace{-3mm}

%---------------------------------------------------------------------
\subsection{Context-Aware Reinforcement Learning}
\label{sec:contrastive_rl}
%---------------------------------------------------------------------

We use GRPO as the base RL objective. In the agentic setting, rewards are binary based on whether the generated patch passes all test cases. In the multimodal setting, the reward is based on exact answer match. However, these rewards do not distinguish whether correct outputs are grounded in the provided context. We therefore add a \emph{context-awareness loss}.

\vspace{1ex}\noindent\textbf{Context-awareness loss.} Given a contrastive instance $z=(Q,A,C^+,C^-)$, the model should assign higher confidence to the positive context $C^+$ over the negative context $C^-$. We operationalize this as a two-way multiple choice: a single prompt presents $Q$, $A$, and the two contexts as labeled options (``A''/``B''), with the order of $C^+$ and $C^-$ randomized per instance to remove position bias (prompt templates in Appendix~\ref{app:prompt_image_choose},~\ref{app:prompt_traj_choose}). Let $t^+$ and $t^-$ be the option-letter tokens assigned to $C^+$ and $C^-$, and let $\ell_\theta^+(z)$ and $\ell_\theta^-(z)$ be the model's next-token logits for $t^+$ and $t^-$ at the answer position, computed by teacher forcing rather than from a sampled rollout. We define the margin $\Delta_\theta(z) = \ell_\theta^+(z) - \ell_\theta^-(z)$ and optimize:
\begin{equation}
    \mathcal{L}_{\mathrm{CA}}(z;\theta)
    =
    -\log
    \sigma
    \left(
    \mathrm{clip}
    \left(
    \Delta_\theta(z),
    -c,
    c
    \right)
    \right),
    \label{eq:ca_loss}
\end{equation}
where $\sigma(\cdot)$ is the sigmoid function and $c>0$ controls margin clipping. This objective encourages correct context selection while preventing large margins from dominating training. Importantly, the loss is modality-agnostic and applies uniformly to agent trajectories and images.

\vspace{1ex}\noindent\textbf{Joint objective.}
Let $\mathcal{D}_{\mathrm{RL}}$ denote the task data and $\mathcal{D}_{\mathrm{CA}}$ the contrastive dataset. We optimize:
\begin{equation}
    \mathcal{L}(\theta)
    =
    \mathbb{E}_{x \sim \mathcal{D}_{\mathrm{RL}}}
    \left[
    \mathcal{L}_{\mathrm{GRPO}}(x;\theta)
    \right]
    +
    \lambda
    \mathbb{E}_{z \sim \mathcal{D}_{\mathrm{CA}}}
    \left[
    \mathcal{L}_{\mathrm{CA}}(z;\theta)
    \right],
    \label{eq:combined}
\end{equation}
where $\lambda>0$ balances task performance and context awareness. The two objectives are complementary. GRPO optimizes for correct outputs, while $\mathcal{L}_{\mathrm{CA}}$ enforces alignment between outputs and their supporting context.
We apply the same formulation across both agentic and multimodal settings.

%---------------------------------------------------------------------
\section{Long Horizon Experiments}
\label{sec:exp_agentic}

In this section, we evaluate \ours~in the long-horizon setting using our constructed agentic contrast context pairs. We first describe the experimental setup, including the base models and baselines (\S\ref{sec:agentic_setup}). Then, we present a comprehensive evaluation across five agentic and long-context benchmarks (\S\ref{sec:agentic_main}). We demonstrate that our method consistently outperforms the RL baseline on \textbf{every} benchmark and exhibits strong \textbf{generalization} beyond the training distribution.
%---------------------------------------------------------------------

\subsection{Experimental Setup}
\label{sec:agentic_setup}

\vspace{1ex}\noindent\textbf{Base models.} We experiment with two base models with different levels of agentic coding capability: Qwen3-8B~\citep{yang2025qwen3technicalreport}, a general-purpose model, and Klear-AgentForge-8B~\citep{wang2025klearagentforgeforgingagenticintelligence}, a model specifically fine-tuned for complex agentic coding.

\vspace{1ex}\noindent\textbf{Baselines.} We compare three training configurations per base model: (i) Base (no RL), (ii) RL baseline (standard GRPO), and (iii) \ours. Detailed training configurations can be found in Appendix~\ref{app:hp_agentic}. In total, our agentic training set comprises 8k instances: 7k standard SWE-Gym~\citep{pan2025trainingsoftwareengineeringagents} and SWE-Smith~\citep{yang2025swesmithscalingdatasoftware} coding tasks optimized with $\mathcal{L}_{\mathrm{GRPO}}$ and 1k contrastive trajectory pairs used for $\mathcal{L}_{\mathrm{CA}}$; to match the total data budget, the RL baseline is trained on 8k tasks from the same source.
We additionally report \emph{off-the-shelf} reference models at larger scales, including Qwen3-14B, Qwen3-32B~\citep{yang2025qwen3technicalreport}, and Qwen3-Coder-30B~\citep{yang2025qwen3technicalreport}.

\vspace{1ex}\noindent\textbf{Evaluation benchmarks.} We evaluate on two in-distribution (ID) and three out-of-distribution (OOD) benchmarks. The in-distribution suite targets agentic coding via SWE-Bench Verified~\citep{yang2024sweagent} and SWE-Bench Lite~\citep{yang2024sweagent}, with performance measured by resolve rate (\%). To assess whether the gains from \ours\ generalize beyond agentic coding, we further evaluate on three OOD benchmarks spanning diverse forms of context utilization: LiveCodeBench v6~\citep{jain2024livecodebenchholisticcontaminationfree} for competitive-programming (solve rate \%), LongBench v2~\citep{bai2024longbench2} for long-context QA (accuracy \%), and Needle-in-a-Haystack (NIAH)~\citep{kamradt2023niah} for targeted retrieval. For NIAH, we report the mean recall (\%) averaged across all needle depths and context lengths.

\subsection{Main results}
\label{sec:agentic_main}

Table~\ref{tab:agentic_main} reports the performance (\%) of all baselines and our trained models across five benchmarks.

\begin{table}[t]
\centering
\small
\setlength{\tabcolsep}{3pt}
\begin{tabular*}{\linewidth}{@{\extracolsep{\fill}}c | cccccc}
\toprule
& \multicolumn{2}{c}{\textbf{In-distribution}} & \multicolumn{4}{c}{\textbf{Out-of-distribution}} \\
\cmidrule(lr){2-3} \cmidrule(lr){4-7}
\multirow{2}{*}{\textbf{Model}} & \multirow{2}{*}{\shortstack{SWE-Bench\\Verified}} & \multirow{2}{*}{\shortstack{SWE-Bench\\Lite}} & \multirow{2}{*}{\shortstack{LiveCodeBench\\v6}} & \multicolumn{2}{c}{LongBench v2} & \multirow{2}{*}{NIAH} \\
\cmidrule(lr){5-6}
& & & & Overall & Long & \\
\midrule
\rowcolor{secgray}
\multicolumn{7}{l}{\emph{Off-the-shelf reference models}} \\
Qwen3-14B           &  8.40 &  6.00 & 57.1 & 34.2 & 24.1 & 99.5 \\
Qwen3-32B           &  8.40 &  6.00 & 61.1 & 36.8 & 31.5 & 99.3 \\
Qwen3-Coder-30B     & 28.8  & 22.0  & 37.7 & 42.5 & 41.7 & 85.7 \\
\midrule
\rowcolor{secgray}
\multicolumn{7}{l}{\emph{Trained from Qwen3-8B}} \\
Base model          &  5.00 &  2.70 & 44.6 & 31.6 & 27.8 & 98.8 \\
RL baseline         &  6.20 &  2.70 & 46.3 & 31.8 & 26.9 & 98.5 \\
\cellcolor{ourpurple}\textbf{Ours} & \textbf{7.00} & \textbf{4.00} & \textbf{47.4} & \textbf{33.2} & \textbf{29.6} & \textbf{99.0} \\[1pt]
$\Delta$& \dt{+0.8} & \dt{+1.3} & \dt{+1.1} & \dt{+1.4} & \dt{+2.7} & \dt{+0.5} \\
\midrule
\rowcolor{secgray}
\multicolumn{7}{l}{\emph{Trained from Klear-AgentForge-8B}} \\
Base model          & 26.6  & 21.0  & 21.7 & 27.4 & 21.3 & 68.3 \\
RL baseline         & 28.0  & 21.7  & 22.3 & 27.0 & 24.1 & 65.5 \\
\cellcolor{ourpurple}\textbf{Ours} & \textbf{30.2} & \textbf{24.0} & \textbf{24.0} & \textbf{29.6} & \textbf{28.7} & \textbf{71.3} \\[1pt]
$\Delta$ & \dt{+2.2} & \dt{+2.3} & \dt{+1.7} & \dt{+2.6} & \dt{+4.6} & \dt{+5.8} \\
\bottomrule
\end{tabular*}
\vspace{0.3em}
\caption{\textbf{Main results on 5 long-horizon benchmarks.} \ours\ consistently outperforms RL baseline across all the tasks for both base models. Notably, \ours\ demonstrates robust generalization to long-context OOD tasks where standard outcome-based RL struggles or regresses.}
\label{tab:agentic_main}
\vspace{-3em}
\end{table}

% \vspace{1ex}\noindent\textbf{Analysis.}

\textbf{Consistent gains across base models.} \ours~improves over the RL baseline on \emph{every} benchmark and for both base models. This highlights that the context-awareness objective (Eq.~\ref{eq:combined}) provides a robust training signal beyond outcome-only RL.

\textbf{Competitive with much larger reference models.} When trained from Klear-AgentForge-8B, \ours\ substantially outperforms Qwen3-32B (4$\times$ larger) and the code-specialized Qwen3-Coder-30B on SWE-Bench. This shows that, in the agentic-coding setting, a well-targeted context-aware objective combined with a domain-adapted base can compensate for substantial differences in scale and pretraining specialization.

\textbf{Generalization to out-of-distribution evaluations.} On \textbf{every} OOD benchmark and on both base models, \ours\ outperforms the base model and the RL baseline. On long-context tasks (LongBench v2, NIAH), the effect is particularly pronounced: On NIAH, standard outcome-based GRPO \emph{regresses} relative to the base model, whereas \ours\ surpasses the base in both cases. On LongBench v2, \ours\ improves the base models both on the overall set and the long subset, with the gain especially pronounced on the long-input subset where input length stretches context grounding the furthest. This pattern indicates that the learned signal transfers beyond the agentic training domain and captures a more general notion of context grounding.

\textbf{Ablations.} 
We investigate the effect of the context-awareness loss coefficient $\lambda$ and find that performance peaks at $\lambda=0.005$, which balances context-awareness and task optimization. Details are in Appendix~\ref{app:ablation_agentic}.

%=====================================================================
\section{Multimodal Experiments}
\label{sec:exp_multimodal}
%=====================================================================

We further evaluate \ours~in the multimodal setting, where correct predictions require tight coupling between visual perception and downstream reasoning. 
We first describe the experimental setup and baselines (\S\ref{sec:setup}), and then present results across 12 diverse benchmarks (\S\ref{sec:main_results}). 
Across both base models and all evaluation categories, \ours~consistently improves performance, indicating stronger visual grounding that propagates to higher-level reasoning.

\subsection{Experimental Setup}
\label{sec:setup}

\vspace{1ex}\noindent\textbf{Base models.} We experiment with two state-of-the-art vision-language models of comparable scale: Qwen2.5-VL-7B-Instruct~\citep{bai2025qwen25vltechnicalreport} and Qwen3-VL-8B-Instruct~\citep{bai2025qwen3vltechnicalreport}.

\vspace{1ex}\noindent\textbf{Baselines.} We focus on \emph{training configurations} to isolate the effect of the learning objective.
For each base model, we compare: (i) the base model without additional training, (ii) the RL baseline trained with standard GRPO, and (iii) Ours, which augments GRPO with the proposed context-awareness loss. Detailed hyper-parameters and training protocols are provided in Appendix~\ref{app:hp_multimodal}. In total, our multimodal training set comprises 45k examples: 38k standard single-image task instances optimized with $\mathcal{L}_{\mathrm{GRPO}}$ and 7k contrastive image pairs used for $\mathcal{L}_{\mathrm{CA}}$; to match the total data budget, the RL baseline is trained on 45k single-image QA examples from the same source. For Qwen2.5-VL, we additionally include (iv) PAPO~\citep{wang2026perceptionawarepolicyoptimizationmultimodal}, a recent method that improves multimodal RL by introducing perception-aware reward shaping, as a baseline. Notably, PAPO is trained on its own curated dataset and uses a different RL formulation, making it a strong but non-directly comparable reference point. Including it helps contextualize the magnitude of gains achievable through alternative approaches.

\vspace{1ex}\noindent\textbf{Evaluation benchmarks.} We evaluate on 12 diverse multimodal benchmarks spanning mathematical reasoning (MathVista~\citep{lu2024mathvista}, MathVerse~\citep{zhang2024mathverse}, MathVision~\citep{wang2024measuringmultimodalmathematicalreasoning}), general multimodal understanding (MMMU-Pro~\citep{yue2024mmmupro}, MMMU~\citep{yue2024mmmu}), fine-grained visual perception (V*~\citep{wu2024vstar}, MMStar~\citep{chen2024mmstar}, BLINK~\citep{fu2024blink}), scientific reasoning (ScienceQA~\citep{lu2022scienceqa}, PhyX~\citep{xue2024phyx}, OlympiadBench Physics~\citep{he2024olympiadbench}), and real-world scene understanding (MME-RealWorld Lite~\citep{xia2024mmrealworld}). For consistency and reproducibility, all evaluations are conducted with two widely-adopted toolkits, LMMs-Eval~\citep{zhang2024lmmsevalrealitycheckevaluation} and VLMEvalKit~\citep{duan2024vlmevalkit}, which yield scores that most closely match those reported in prior work.

\subsection{Main Results}
\label{sec:main_results}

Table~\ref{tab:main_results} presents results across both base models and all 12 benchmarks.

\begin{table}[t]
\centering
\small
\setlength{\tabcolsep}{3pt}
\begin{tabular*}{\linewidth}{@{\extracolsep{\fill}}c | ccc >{}c >{}c c cc >{}c >{}c}
\toprule
& \multicolumn{5}{c}{\textbf{Qwen2.5-VL-7B}} & & \multicolumn{4}{c}{\textbf{Qwen3-VL-8B}} \\
\cmidrule(lr){2-6} \cmidrule(lr){8-11}
\textbf{Benchmark} & Base & RL Base & PAPO & \cellcolor{ourpurple}\textbf{Ours} & $\Delta$ & & Base & RL Base & \cellcolor{ourpurple}\textbf{Ours} & $\Delta$ \\
\midrule
\rowcolor{secgray}
\multicolumn{11}{l}{\emph{Mathematical Reasoning}} \\
\quad MathVista          & 68.2 & 72.5 & 72.7 & \textbf{73.6} & \dt{+1.1} & & 75.8 & 78.7 & \textbf{79.8} & \dt{+1.1} \\
\quad MathVerse          & 43.9 & 45.3 & \textbf{49.7} & 49.1 & \dt{+3.8} & & 56.1 & 65.0 & \textbf{66.4} & \dt{+1.4} \\
\quad MathVision         & 22.8 & 25.5 & \textbf{27.3} & 26.8 & \dt{+1.3} & & 46.2 & 49.2 & \textbf{52.0} & \dt{+2.8} \\
\midrule
\rowcolor{secgray}
\multicolumn{11}{l}{\emph{General Multimodal Understanding}} \\
\quad MMMU-Pro           & 36.6 & 41.3 & 42.6 & \textbf{42.8} & \dt{+1.5} & & 41.3 & 55.9 & \textbf{57.5} & \dt{+1.6} \\
\quad MMMU               & 50.7 & 53.3 & 53.2 & \textbf{54.6} & \dt{+1.3} & & 66.4 & 69.1 & \textbf{70.1} & \dt{+1.0} \\
\midrule
\rowcolor{secgray}
\multicolumn{11}{l}{\emph{Fine-grained Visual Perception}} \\
\quad V* & 70.1 & 70.7 & 71.7 & \textbf{73.3} & \dt{+2.6} & & 82.2 & 84.8 & \textbf{85.9} & \dt{+1.1} \\
\quad MMStar             & 62.6 & 64.1 & 63.4 & \textbf{65.1} & \dt{+1.0} & & 70.5 & 73.5 & \textbf{74.8} & \dt{+1.3} \\
\quad BLINK              & 55.3 & 56.5 & 58.5 & \textbf{58.9} & \dt{+2.4} & & 64.4 & 65.1 & \textbf{66.6} & \dt{+1.5} \\
\midrule
\rowcolor{secgray}
\multicolumn{11}{l}{\emph{Scientific Reasoning}} \\
\quad ScienceQA          & 88.2 & 91.0 & 92.7 & \textbf{95.4} & \dt{+4.4} & & 94.4 & 95.6 & \textbf{96.6} & \dt{+1.0} \\
\quad PhyX               & 25.4 & 48.7 & 46.8 & \textbf{50.0} & \dt{+1.3} & & 45.5 & 72.1 & \textbf{73.4} & \dt{+1.3} \\
\quad OlympiadBench Phy  &  1.5 &  3.1 &  2.2 & \textbf{4.6}  & \dt{+1.5} & &  7.9 &  8.1 & \textbf{9.9} & \dt{+1.8} \\
\midrule
\rowcolor{secgray}
\multicolumn{11}{l}{\emph{Real-world Scene Understanding}} \\
\quad MME-RealWorld Lite & 38.4 & 45.1 & 45.1 & \textbf{46.7} & \dt{+1.6} & & 48.7 & 51.9 & \textbf{54.8} & \dt{+2.9} \\
\midrule
\textbf{Overall Avg.}    & 47.0 & 51.4 & 52.2 & \textbf{53.4} & \dt{+2.0} & & 58.3 & 64.1 & \textbf{65.7} & \dt{+1.6} \\
\bottomrule
\end{tabular*}
\vspace{0.3em}
\caption{\textbf{Main results on 12 diverse multimodal benchmarks.} \ours\ achieves the best performance across all sub-categories, surpassing both the RL baseline and strong reference methods like PAPO. The broad improvements indicate that \ours\ successfully translates enhanced visual grounding into stronger end-to-end reasoning capabilities.}
\label{tab:main_results}
\vspace{-2.5em}
\end{table}

\textbf{Consistent gains across benchmarks and categories.} \ours~outperforms the RL baseline on \textbf{every} benchmark for both base models, and across all five sub-task categories. The improvements are not concentrated in any single regime, but instead span perception-heavy (V*, MMStar, BLINK, MME-RealWorld Lite), reasoning-heavy (MathVista, MathVerse, MathVision, PhyX, OlympiadBench Physics), and mixed tasks, with no category trading off against another. This breadth makes it unlikely that the context-awareness loss is merely shifting capability between categories. Rather, the consistency across categories and across two model families suggests the gains reflect improved context grounding rather than category- or model-specific tuning. To put the magnitude in context, on Qwen2.5-VL, \ours\ improves over the RL baseline by $+2.0$ on average: more than the $+0.8$ of PAPO, a method purpose-built for multimodal perception. The gains are therefore substantial relative to dedicated perception-oriented RL rather than incremental.

\textbf{Ablations.} 
We also investigate the effect of the loss weight $\lambda$ in multimodal setting and find that optimal performance is achieved at small values ($0.001$ or $0.005$, depending on the model). Larger values degrade performance as the auxiliary context- awareness loss begins to overwhelm the primary GRPO objective. Details can be found in Appendix~\ref{app:ablation_multimodal}.

\section{Comparison between Data Augmentation and \ours}
\label{sec:vs_data_aug}
We investigate whether the gains from \ours~can be reproduced by standard data augmentation using the same contrastive data. This comparison isolates whether the improvement arises from the \emph{data itself} or from the \emph{training objective used to consume it}.

First, we introduce two natural data-augmentation baselines, DA-SFT and DA-RL, which utilize the same contrastive data (\S\ref{sec:da_baselines}).
We then compare their end-task performance in both agentic and multimodal settings (\S\ref{sec:da_results}), demonstrating that standard augmentation either provides no meaningful signal or leads to catastrophic policy collapse (\S\ref{sec:da_results}). Finally, we provide a mechanism study to analyze the effectiveness of \ours (\S\ref{sec:da_mechanism}).

\subsection{Data Augmentation Baselines}
\label{sec:da_baselines}

We evaluate two standard ways of incorporating contrastive data into training: \textbf{DA-SFT}: We first perform supervised fine-tuning on the contrastive data using a cross-entropy objective to predict the correct context. The resulting model is then further optimized with standard GRPO on the task data.
\textbf{DA-RL}: We directly mix contrastive examples into the RL training stream. Each contrastive instance is treated as a binary decision problem, with reward $1$ if the model selects the correct context and $0$ otherwise. Under this formulation, contrastive and task examples share the same outcome-based reward signal.
Both baselines use exactly the same contrastive dataset as \ours\ and are trained for the same number of steps, ensuring a controlled comparison. We evaluate both approaches in the agentic and multimodal settings.

\subsection{End-Task Performance Comparison}
\label{sec:da_results}

\begin{wraptable}[10]{r}{0.65\linewidth}
\centering
\small
\setlength{\tabcolsep}{4pt}
\begin{tabular}{c | cccc}
\toprule
 & \multicolumn{2}{c}{\textbf{Klear-AgentForge-8B}} & \multicolumn{2}{c}{\textbf{Qwen3-8B}} \\
\cmidrule(lr){2-3} \cmidrule(lr){4-5}
\textbf{Configuration} & \textbf{Verified} & \textbf{Lite} & \textbf{Verified} & \textbf{Lite} \\
\midrule
RL Baseline                                & 28.0          & 21.7          & 6.20          & 2.70 \\
DA-SFT                                     &  6.4          &  1.3          & 0.00          & 0.00 \\
DA-RL                                      & 27.6          & 21.7          & 5.60          & 3.00 \\
\cellcolor{ourpurple} \textbf{Ours}                  & \textbf{30.2} & \textbf{24.0} & \textbf{7.00} & \textbf{4.00} \\
\bottomrule
\end{tabular}
\vspace{-1.5mm}

\caption{\textbf{Data augmentation results on agentic benchmarks.}}
\label{tab:da_agentic}
\end{wraptable}
\textbf{Agentic results.} Table~\ref{tab:da_agentic} reports SWE-Bench Verified and SWE-Bench Lite resolve rates for Klear-AgentForge-8B and Qwen3-8B. Two clear patterns emerge across both base models. First, DA-SFT leads to \textbf{catastrophic policy collapse}: on Klear-AgentForge-8B, performance drops from 28.0 / 21.7 under the RL baseline to 6.4 / 1.3, and on Qwen3-8B the model collapses to 0.00 / 0.00. Although the model learns the contrastive task, supervised training on short selection examples severely disrupts the long-horizon interaction policy required for agentic coding. Second, DA-RL is \textbf{nearly indistinguishable} from the RL baseline on both models, indicating that simply mixing contrastive examples into outcome-only RL contributes little additional learning signal. In contrast, \ours\ substantially outperforms both augmentation strategies on both benchmarks and for both base models.

\begin{table}[t]
\centering
\small
\setlength{\tabcolsep}{3pt}
\begin{tabular*}{\linewidth}{@{\extracolsep{\fill}}c | ccc>{}c c ccc>{}c}
\toprule
& \multicolumn{4}{c}{\textbf{Qwen2.5-VL-7B}} & & \multicolumn{4}{c}{\textbf{Qwen3-VL-8B}} \\
\cmidrule{2-5} \cmidrule{7-10}
\textbf{Benchmark} & RL Base & DA-SFT & DA-RL & \cellcolor{ourpurple}\textbf{Ours} & & RL Base & DA-SFT & DA-RL & \cellcolor{ourpurple}\textbf{Ours} \\
\midrule
\quad MathVista          & 72.5 & 70.8 & 73.4 & \textbf{73.6} & & 78.7 & 79.0 & 78.3 & \textbf{79.8} \\
\quad MathVerse          & 45.3 & 47.8 & 45.8 & \textbf{49.1} & & 65.0 & 65.1 & 65.2 & \textbf{66.4} \\
\quad MathVision         & 25.5 & 25.1 & 26.5 & \textbf{26.8} & & 49.2 & 47.3 & 49.5 & \textbf{52.0} \\
\quad MMMU-Pro           & 41.3 & 42.5 & 41.0 & \textbf{42.8} & & 55.9 & 55.3 & 56.2 & \textbf{57.5} \\
\quad MMMU               & 53.3 & 53.7 & 54.1 & \textbf{54.6} & & 69.1 & 68.2 & 69.4 & \textbf{70.1} \\
\quad V*                 & 70.7 & 69.6 & 71.2 & \textbf{73.3} & & 84.8 & 85.3 & 84.8 & \textbf{85.9} \\
\quad MMStar             & 64.1 & 64.0 & 64.0 & \textbf{65.1} & & 73.5 & 73.1 & 73.6 & \textbf{74.8} \\
\quad BLINK              & 56.5 & 55.1 & 57.5 & \textbf{58.9} & & 65.1 & 65.3 & 65.6 & \textbf{66.6} \\
\quad ScienceQA          & 91.0 & 92.4 & 92.1 & \textbf{95.4} & & 95.6 & 95.4 & 95.8 & \textbf{96.6} \\
\quad PhyX               & 48.7 & 49.0 & 48.9 & \textbf{50.0} & & 72.1 & 71.9 & 72.1 & \textbf{73.4} \\
\quad OlympiadBench Phy  &  3.1 &  3.1 &  3.1 & \textbf{4.6}  & &  8.1 &  7.9 &  8.6 & \textbf{9.9} \\
\quad MME-RealWorld Lite & 45.1 & 45.0 & 44.0 & \textbf{46.7} & & 51.9 & 52.8 & 52.8 & \textbf{54.8} \\
\midrule
\textbf{Overall Avg.}    & 51.4 & 51.5 & 51.8 & \textbf{53.4} & & 64.1 & 63.9 & 64.3 & \textbf{65.7} \\
\bottomrule
\end{tabular*}
\vspace{0.3em}
\caption{\textbf{Comparison with standard data augmentation in the multimodal setting.} DA-SFT and DA-RL yield negligible improvements over the RL baseline. In contrast, \ours\ effectively extracts the learning signal to boost performance across all 12 benchmarks.}
\label{tab:da_multimodal}
\vspace{-2.5em}
\end{table}

\vspace{1ex}\noindent\textbf{Multimodal results.}
Table~\ref{tab:da_multimodal} reports results across all 12 multimodal benchmarks for Qwen2.5-VL-7B and Qwen3-VL-8B. Unlike the agentic setting, DA-SFT does not catastrophically fail, but both augmentation strategies remain \textbf{largely ineffective}. DA-SFT achieves averages of 51.5 on Qwen2.5-VL and 63.9 on Qwen3-VL, compared to 51.4 and 64.1 for the RL baseline, respectively. DA-RL yields only marginal gains (+0.4 / +0.2). In contrast, \ours\ improves average performance by +2.0 points on Qwen2.5-VL and +1.6 points on Qwen3-VL, while improving \emph{every} benchmark individually. These results suggest that the contrastive data alone is insufficient: the gains arise specifically from the way the signal is incorporated into training.

\subsection{Mechanism Analysis: Why Data Augmentation Fails}
\label{sec:da_mechanism}

\begin{figure}[t]
\centering
\includegraphics[width=\linewidth]{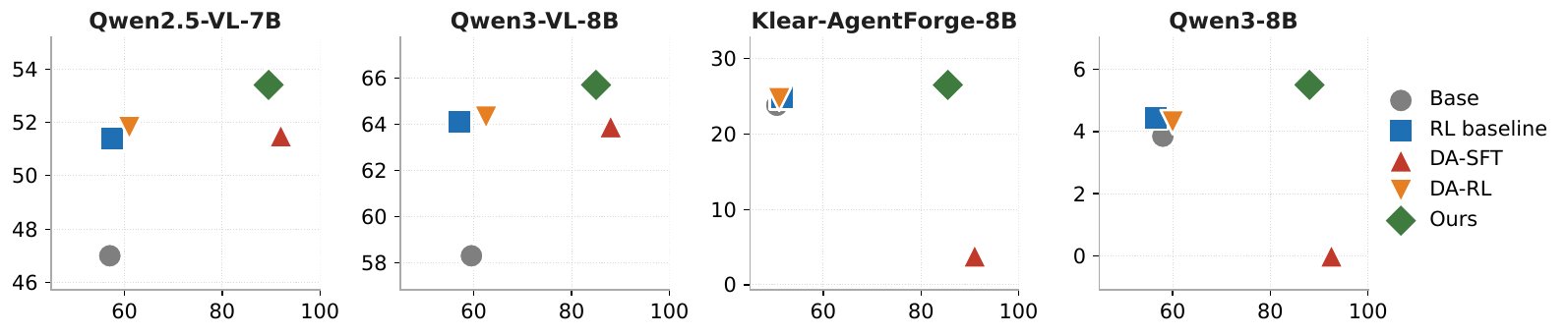}
\caption{\textbf{Context-selection accuracy versus end-task performance.} The $x$-axis denotes selection accuracy. The $y$-axis denotes end-task performance. Top-right is optimal. DA-SFT achieves high selection accuracy but collapses on the $y$-axis. DA-RL fails to learn discrimination at all. \ours\ is the \emph{only} method that couples high context awareness with end-task improvements.}
\label{fig:disc_acc_post}
\vspace{-2.5em}
\end{figure}

\vspace{1ex}\noindent\textbf{Mechanism check: selection accuracy vs.\ end-task performance.}
The end-task comparisons above establish \emph{which} methods help, but not \emph{why}. To test whether each method learns context selection, we evaluate every method on the held-out test set introduced in Fig.~\ref{fig:choice}, and plot selection accuracy against end-task performance in Fig.~\ref{fig:disc_acc_post}. Three patterns emerge. 

\emph{(i) Outcome-only RL fails to learn context selection.} The RL baseline stays near the base-model cluster across all panels, showing that final-task rewards alone provide little signal for distinguishing context-grounded reasoning from shortcuts. DA-RL inherits this limitation, since framing context selection as binary reward prediction merely repackages the same sparse outcome signal. 

\emph{(ii) DA-SFT learns context selection but harms the policy.} DA-SFT achieves the highest selection accuracy in nearly every setting, confirming a strong supervision signal in the contrastive data. However, this rarely translates to downstream performance gains. On both agentic base models, it nearly collapses end-task performance. We attribute this to SFT on short-answer examples shifting the model distribution away from the long-form, multi-turn patterns the agent scaffold requires. While this mismatch is mild for single-turn multimodal responses, it is catastrophic in long-horizon agentic settings. 

\emph{(iii) Context selection skill alone is insufficient.} Both DA-SFT and \ours\ push selection accuracy to 85--93\%, but only \ours\ consistently improves downstream task performance. This suggests that context selection is necessary but not sufficient: the model must acquire this capability without disrupting the policy behaviors required for the original task distribution. This also speaks directly to the artifact concern: if context selection were being solved by exploiting construction artifacts, the configuration with the \emph{highest} selection accuracy (DA-SFT) should transfer best; instead it transfers worst. Whatever signal drives high selection accuracy---including any residual artifact---is therefore, on its own, insufficient to explain the end-task gains of \ours.

\vspace{1ex}\noindent\textbf{Why \ours\ avoids both failure modes.}
\ours\ leverages the supervision in contrastive data without the instability of direct augmentation, by adding the context-awareness loss as a bounded auxiliary objective inside the on-policy GRPO loop. Two properties are critical.

\emph{(a) Updates remain constrained.} The GRPO importance-ratio clipping term $\mathrm{clip}(\rho_t^{(i)}, 1{-}\epsilon, 1{+}\epsilon)$ and KL regularization keep the policy close to the reference model $\pi_{\mathrm{ref}}$, while the clipped margin objective $\mathrm{clip}(\Delta_\theta(z), -c, c)$ in Eq.~\ref{eq:ca_loss} suppresses auxiliary gradients once $C^+$ and $C^-$ are well separated. Together, these mechanisms preserve the original policy distribution and prevent the catastrophic forgetting observed in DA-SFT.

\emph{(b) The auxiliary signal is dense.} Unlike DA-RL's sparse ${0,1}$ rewards, $\mathcal{L}_{\mathrm{CA}}$ directly supervises the relative preference between $C^+$ and $C^-$ on every example. Meaningful gradients are therefore produced even when the current policy would rarely sample the correct context, precisely the regime where outcome-only RL struggles. Figure~\ref{fig:disc_acc_post} reflects this clearly: \ours\ is the only method that both achieves high context selection accuracy and strong end-task performance across all settings.

Overall, the contrastive data alone is insufficient. The key is the objective used to consume it.

\vspace{-2mm}
\section{Related Work}
\vspace{-2mm}
\label{sec:relatedworks}

\vspace{1ex}\noindent\textbf{RL Post-training in Agentic and Multimodal Settings.}
Reinforcement learning has become a standard recipe for post-training LLMs. Building on RLHF~\citep{ouyang2022traininglanguagemodelsfollow} and DPO~\citep{rafailov2024directpreferenceoptimizationlanguage}, recent verifiable-reward methods such as GRPO~\citep{shao2024deepseekmath} and DAPO~\citep{yu2025dapoopensourcellmreinforcement} have driven rapid progress on mathematical and coding reasoning~\citep{Guo_2025, wei2025swerladvancingllmreasoning}. The same recipe has been adapted to the agentic setting, where SWE-RL~\citep{wei2025swerladvancingllmreasoning} and DeepSWE~\citep{deepswe2025} explore RL as a new training paradigm for software-engineering agents. It also extends naturally to multimodal LLMs~\citep{bai2025qwen25vltechnicalreport, bai2025qwen3vltechnicalreport}, where Vision-R1~\citep{huang2026visionr1incentivizingreasoningcapability} and PAPO~\citep{wang2026perceptionawarepolicyoptimizationmultimodal} refine the RL pipeline for visual reasoning, cold-start pipelines pair SFT with RL~\citep{wei2025coldstart}, and consistency-aware rewards extend GRPO~\citep{chen2025grpocare}. RL has likewise become central to multimodal \emph{agents} that interleave reasoning with tool use and search~\citep{chng2025sensenovamars}, reason over hour-long videos~\citep{liu2025longvideoagent, zhong2025omnir1, zhou2025reinforcedmllm}. Most of these methods optimize outcome correctness or answer--rationale consistency, yet none supplies a signal about whether the model actually grounds its answer in the supplied context. Our method \ours\ targets this gap: it is a modality-agnostic training method, compatible in principle with standard policy-gradient algorithms, and supplies a process-level signal that complements outcome rewards. The concurrent work ContextRL of Lu~\etal~\citep{lu2026contextrl} also couples context with multimodal RL, but along an orthogonal axis: it feeds reference solutions to the \emph{reward model} as additional context for fine-grained process verification. In contrast, \ours\ makes context the \emph{object} of an auxiliary selection task for the policy itself and applies uniformly across agentic and multimodal settings.

\vspace{1ex}\noindent\textbf{Context Utilization and Contrastive Supervision.}
A growing body of work~\citep{liu2023lostmiddlelanguagemodels, hsieh2024rulerwhatsrealcontext, mei2025surveycontextengineeringlarge} shows that LLMs often fail to use their supplied context faithfully. Two distinct lines of work attempt to address this. The first targets long-context settings directly: FILM~\citep{an2024makellmfullyutilize} introduces information-intensive supervision for long-context retrieval, LongRLVR~\citep{chen2026longrlvrlongcontextreinforcementlearning} adds context rewards in long-context RL, and MemOCR~\citep{shi2026memocr} preserves sparse but decisive evidence through memory- and layout-aware compression rather than contrastive selection. The second draws on contrastive supervision, widely adopted in supervised fine-tuning and DPO-style objectives: VC-STaR~\citep{pan2026lenscontrastselfimprovingvisual} improves VLM reasoning by contrasting visually similar VQA pairs, mDPO~\citep{wang2024mdpoconditionalpreferenceoptimization} augments DPO~\citep{rafailov2024directpreferenceoptimizationlanguage} with an image-side preference term, MMEmb-R1~\citep{wang2026mmembr1} performs pair-aware contrastive selection for multimodal embedding, and CARE~\citep{wang2025carewhatfails} contrasts a correct rollout against hard-negative failures for verifiable reasoning. \ours \, shares this contrastive spirit but operates along a different axis: rather than preferring one response over another under a fixed context, we prefer one context over another under a fixed \textit{(Query, Answer)} pair.

\vspace{-2mm}
\section{Conclusion}
\vspace{-2mm}
\label{sec:conclusion}

We introduce \ours, a context-aware RL framework that adds a lightweight context-selection objective to outcome-based post-training. Across 17 benchmarks, \ours~improves over GRPO in both long-horizon and multimodal settings. Comparisons with data-augmentation baselines show that the gains come from the training objective rather than the contrastive data alone. These results position context selection as a simple and consistently beneficial auxiliary signal for improving context grounding.

\newpage
\bibliographystyle{plain}
\bibliography{neurips_2026}

\newpage
\appendix
\section{Additional Method Details}

\subsection{Detailed Procedure for Mining Agentic Contrast Trajectories}
\label{app:agentic_mining}

We mine contrastive trajectory pairs from the 66k SWE-smith trajectories~\citep{wang2025klearagentforgeforgingagenticintelligence,yang2025swesmithscalingdatasoftware} with the following step-by-step procedure.

\begin{enumerate}
    \item \emph{Group by repository and commit.}
    We group trajectories by repository and commit. This ensures that paired trajectories share the same underlying code base.

    \item \emph{Filter by modified file.}
    Within each group, we keep only pairs whose reference patches modify the same file. This makes the two contexts comparable at the file level.

    \item \emph{Filter by target function or class.}
    We parse the patch hunks and identify the function or class touched by each patch. We keep pairs that modify the same function or class. This creates pairs that differ in a small but important region of the code.

    \item \emph{Filter by issue relation.}
    We keep pairs whose issue descriptions are distinct but related. For example, two issues may refer to the same API, the same corner case family, or the same function behavior. This step avoids redundant pairs while preserving semantic relatedness.

    \item \emph{Mask direct patch leakage.}
    We replace patch content inside edit commands with \texttt{<PATCH\_MASKED>}. This prevents the model from solving the context selection task by directly reading the edit command. We do not mask file views, test outputs, error messages, or reasoning traces, because these are valid parts of the context available to the agent.
\end{enumerate}

Each of these filters is applied conservatively, so the number of qualifying pairs drops sharply at every stage. We further verify all resulting pairs with GPT~5.4 followed by manual inspection, removing pairs that remain ambiguous or contain residual leakage. After this pipeline, only 1k pairs out of the original 66k trajectories survive, forming our final agentic contrastive training set.

%---------------------------------------------------------------------
\subsection{Multimodal Data Sources}
\label{app:multimodal_sources}
%---------------------------------------------------------------------

We utilize source datasets spanning five visual domains to construct multimodal contrast context pairs. We use generative editing for natural images and similarity-based retrieval for structured visual inputs. The source datasets used for each domain are as follows:

\begin{itemize}
    \item \textbf{Chart:} ChartQA, DVQA
    \item \textbf{Geometry:} Geo170K, Geometry3K, GeomVerse, GeoQA3, MAVIS-Geometry, R-CoT
    \item \textbf{Non-geometric math:} ICON-QA, K12, MAVIS-Function, MMK12, MM-Math
    \item \textbf{Science:} AI2D, M3CoT, ScienceQA
    \item \textbf{Natural images:} Visual CoT
\end{itemize}

%---------------------------------------------------------------------
\subsection{Dataset Difficulty and Quality Control: Screening for Artifacts and Shortcut Cues}
\label{app:data_quality}
%---------------------------------------------------------------------

A natural concern for any newly constructed contrastive dataset is whether the positive context can be selected by exploiting construction artifacts or superficial shortcut cues rather than by genuine context grounding. We address this along three axes: (i) the verifiers used to build the data are themselves explicit artifact/shortcut filters; (ii) the pairs are hard by construction; and (iii) the downstream evidence (OOD transfer and the data-augmentation controls) is inconsistent with artifact exploitation.

\paragraph{Quality control is an artifact/shortcut filter.}
Both verifiers are designed to \emph{reject}, rather than tolerate, the cues a shortcut solver would exploit. For trajectories, criterion~(iii) of the verifier rejects pairs with large length or formatting disparities, patch-specific tokens that leak into only one trajectory, only one trajectory inspecting the modified file or function, or masking applied inconsistently between the two trajectories. For images, criterion~(i) (Appendix~\ref{app:prompt_edit_verifier}) rejects edited images with visible editing artifacts (blur, warping, broken boundaries, implausible lighting, texture mismatch, floating objects), and criterion~(iii) rejects global restyling or any change to question-irrelevant regions that could serve as a whole-image ``tell''. In other words, the superficial-cue analysis one might run as a post-hoc audit is built directly into our acceptance test: a pair solvable by such cues is, by design, rejected. Any case the verifier cannot confidently clear is marked \texttt{UNCERTAIN} and escalated to manual review by the authors, who apply the same criteria.

\paragraph{Pairs are hard by construction.}
The negatives are not random; the construction constrains them to lie near the similarity ceiling. \emph{(i) Agentic.} Paired trajectories share the same repository, commit, modified file, and target function or class, and differ only in a small decisive code region, with edit commands masked by \texttt{<PATCH\_MASKED>}; the two contexts are therefore nearly identical at the token level. \emph{(ii) Multimodal, generative editing.} The negative is a localized edit of the positive that changes only the answer-relevant region while preserving the rest of the scene, so the two images are identical almost everywhere. \emph{(iii) Multimodal, retrieval.} Negatives are retrieved under a high cosine-similarity threshold $\cos(f_I(I),f_I(I')) \ge \alpha_I$, so they are visually close to the positive by construction. Selection therefore requires resolving a small, decisive difference rather than a salient one.

\paragraph{Aggressive rejection.}
Each construction pipeline retains only a small, high-precision fraction of candidates after automatic verification and manual review, as summarized in Table~\ref{tab:funnel}.

\begin{table}[ht]
\centering
\small
\setlength{\tabcolsep}{8pt}
\begin{tabular}{l l r r}
\toprule
\textbf{Setting} & \textbf{Construction} & \textbf{Candidates} & \textbf{Retained} \\
\midrule
Agentic    & cascade mining + verify & 66{,}000 traj. & 1{,}000 ($1.5\%$) \\
Multimodal & generative editing      & 2{,}000        & $\sim$700 ($\sim$35\%) \\
Multimodal & similarity retrieval    & $>$200{,}000    & 6{,}300 ($\sim$3.1\%) \\
\bottomrule
\end{tabular}
\caption{\textbf{Filtering funnel.} Each construction pipeline retains only a small, high-precision fraction of candidate pairs after automatic verification and manual review.}
\label{tab:funnel}
\end{table}

\paragraph{Why this is unlikely to be artifact exploitation.}
Two pieces of evidence in the main text bear on whether the learned signal is artifact detection. First, \ours\ improves on benchmarks that contain \emph{none} of our constructed inputs: the agentic OOD suite (LiveCodeBench, LongBench~v2, NIAH) contains no mined pairs or masked patches, and all 12 multimodal benchmarks use natural, unedited single images in the standard (non-selection) format. An artifact-detection skill would be inapplicable on these benchmarks by construction, yet the gains transfer broadly---behavior expected of content-level grounding, not of artifact detection. Second, in the data-augmentation study (\S\ref{sec:da_mechanism}), DA-SFT attains the \emph{highest} context-selection accuracy (85--93\%) yet fails to improve, and often collapses, downstream performance; if high selection accuracy were driven by exploitable artifacts, the configuration that maximizes it should transfer best, whereas the opposite holds. Finally, in the generative-editing pipeline the editor and verifier are distinct models (Nano Banana 2 and GPT~5.4, respectively), so the verifier audits another model's output rather than its own, reducing the chance that generator-specific artifacts pass undetected. Together these indicate that what \ours\ learns from the contrastive pairs is not reducible to construction-specific artifacts.

\section{Dataset Composition}
\label{app:dataset_composition}

This section gives the per-source breakdown of the multimodal training set used in \S\ref{sec:exp_multimodal}. The complete training set contains \textbf{45{,}000} examples drawn from a single \texttt{jsonl} file: \textbf{38{,}000} standard single-image task examples used for the GRPO objective $\mathcal{L}_{\mathrm{GRPO}}$, and \textbf{7{,}000} two-image contrastive instances used for the context-selection objective $\mathcal{L}_{\mathrm{CA}}$. The 7{,}000 contrastive instances correspond to the \textbf{7k contrastive image pairs} reported in \S\ref{sec:data_multimodal}, and the 85\%/15\% task vs.\ contrastive split matches the proportion described in \S\ref{sec:exp_multimodal}.

\begin{table}[ht]
\centering
\small
\setlength{\tabcolsep}{6pt}
\begin{tabular}{l l r r}
\toprule
\textbf{Subgroup} & \textbf{Source dataset} & \textbf{Count} & \textbf{\% of total} \\
\midrule
\multirow{6}{*}{Geometry}
  & MAVIS-Geometry & 18{,}776 & 41.72 \\
  & Geo170K        &  1{,}224 &  2.72 \\
  & R-CoT          &     344 &  0.76 \\
  & Geometry3K     &     140 &  0.31 \\
  & GeoQA3         &     128 &  0.28 \\
  & GeomVerse      &     117 &  0.26 \\
\cmidrule(lr){2-4}
  & \emph{Subtotal} & \textbf{20{,}729} & \textbf{46.06} \\
\midrule
\multirow{3}{*}{Non-geometric Math}
  & MAVIS-Function &  6{,}914 & 15.36 \\
  & ICON-QA        &     763 &  1.70 \\
  & MMK12          &      56 &  0.12 \\
\cmidrule(lr){2-4}
  & \emph{Subtotal} & \textbf{7{,}733} & \textbf{17.18} \\
\midrule
\multirow{2}{*}{Chart}
  & DVQA           &  5{,}485 & 12.19 \\
  & ChartQA        &      40 &  0.09 \\
\cmidrule(lr){2-4}
  & \emph{Subtotal} & \textbf{5{,}525} & \textbf{12.28} \\
\midrule
\multirow{2}{*}{Science}
  & M3CoT          &  3{,}449 &  7.66 \\
  & ScienceQA      &  1{,}634 &  3.63 \\
\cmidrule(lr){2-4}
  & \emph{Subtotal} & \textbf{5{,}083} & \textbf{11.30} \\
\midrule
\multirow{2}{*}{Natural Image}
  & Visual-CoT     &  5{,}600 & 12.44 \\
  & ViRL39K        &     330 &  0.73 \\
\cmidrule(lr){2-4}
  & \emph{Subtotal} & \textbf{5{,}930} & \textbf{13.18} \\
\midrule
\textbf{Total}      &           & \textbf{45{,}000} & \textbf{100.00} \\
\bottomrule
\end{tabular}
\caption{Source-level composition of the 45{,}000-example multimodal training set.}
\label{tab:dataset_composition}
\end{table}

\begin{figure}[htbp]
\centering
\includegraphics[width=0.95\linewidth]{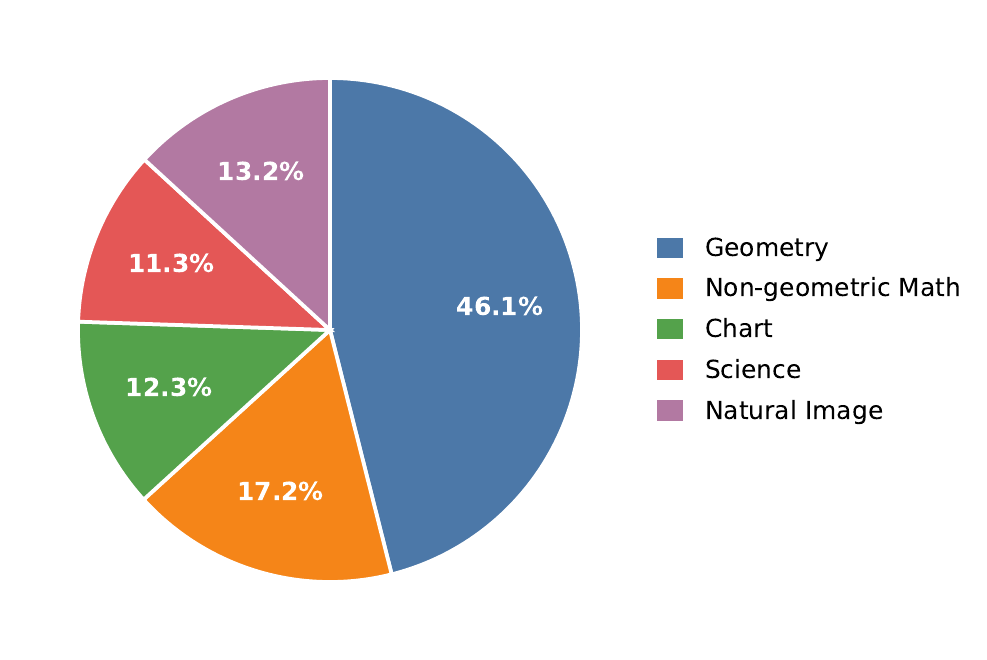}
\caption{\textbf{Subgroup composition of the multimodal training set (45{,}000 examples).} Per-source counts within each subgroup are listed in Table~\ref{tab:dataset_composition}.}
\label{fig:dataset_composition}
\end{figure}

Figure~\ref{fig:dataset_composition} visualizes the subgroup-level composition of the full 45{,}000-example training set across the five visual subgroups (Geometry, Non-geometric Math, Chart, Science, Natural Image). Table~\ref{tab:dataset_composition} reports the source-level breakdown within each subgroup.

Two implementation notes apply to the source-level counts. First, examples in the \emph{Natural Image} subgroup whose answer-relevant content was modified by Nano Banana 2 are saved with new image identifiers; we count both the original and the generatively edited variant under their underlying source dataset (Visual-CoT). Second, in the contrastive subset, both images of a pair share the same source dataset by construction (similarity-based retrieval is performed within a source pool), so every pair contributes a single tally to its source.

\section{Prompt Templates}
\label{app:prompts}

This section lists the prompt templates used in our experiments. Curly braces (e.g.\ \texttt{\{question\}}, \texttt{\{trajectory\_a\}}) denote slots filled with per-instance content; all other text is fixed across instances.

\subsection{Trajectory Selection Prompt}
\label{app:prompt_traj_choose}

This template is used to construct the agentic context-selection instances $z=(Q,A,\tau,\tau')$. The model is shown a code patch and two candidate trajectories whose edit commands have been masked, and must identify which trajectory produced the patch.

\begin{promptbox}{Trajectory Selection Prompt Template}
\begin{Verbatim}[fontsize=\small]

You are given a code patch that fixes a bug in a Python project, along with two
candidate agent trajectories labeled Option A and Option B. Each trajectory shows
an agent exploring a codebase and diagnosing a bug; the actual edit commands in
the trajectories have been replaced with the placeholder <PATCH_MASKED>.

Exactly one of the two trajectories is the one that produced the code patch
shown below. Your task is to identify which trajectory (A or B) produced the
patch.

To decide, reason about:
  - What bug each trajectory is diagnosing (from the agent's reasoning and the
    buggy code it inspects).
  - Whether the given code patch fixes the bug described in trajectory A or in
    trajectory B.

===== CODE PATCH =====
{code_patch}

===== OPTION A =====
{trajectory_a}

===== OPTION B =====
{trajectory_b}

Output only a single letter, either "A" or "B", as your final answer.
\end{Verbatim}
\end{promptbox}

\subsection{Image Selection Prompt}
\label{app:prompt_image_choose}

This template is used to construct the multimodal context-selection instances $z=(Q,A,I,I')$. Two images are placed before the prompt body; the model must select which image is consistent with the reference QA pair.

\begin{promptbox}{Image Selection Prompt Template}
\begin{Verbatim}[breaklines=true,breakanywhere=true,fontsize=\small]
<image>
<image>
Task: You will be presented with two images, a specific question, and its
ground-truth answer. Your goal is to determine which image (if any) accurately
reflects the information provided in the QA pair.

Input Data:
The following QA pair is the reference for your verification:
```text
[REFERENCE DATA]
Question: {question}
Ground-truth Answer: {ground_truth_answer}
```

Select from the following choices.
Options:
A. The first image
B. The second image

Please select the correct answer from the options above.
\end{Verbatim}
\end{promptbox}

\subsection{Contrast Trajectory Pairs Verification Prompt}
\label{app:prompt_traj_verifier}
This is the prompt used by GPT~5.4 to verify the quality of contrast trajectory pairs. The verifier jointly assesses trajectory coherence, answer validity, and the absence of superficial shortcut cues. Examples marked \texttt{UNCERTAIN} are escalated to manual review.
\begin{promptbox}{Trajectory selection Verification Prompt Template}
\begin{Verbatim}[fontsize=\small,breaklines=true,breakanywhere=true]
You are a quality verifier for an agentic trajectory dataset.
You will be shown a code patch together with two candidate agent
trajectories whose edit commands have been masked, and your job is to
decide whether this selection instance is valid for inclusion in
the dataset.

You will receive:
- Code Patch: the diff that fixes a bug in a Python project
- Trajectory A: a candidate agent trajectory with edit commands
  replaced by <PATCH_MASKED>
- Trajectory B: a second candidate trajectory with edit commands
  replaced by <PATCH_MASKED>
- Ground-Truth Label: which trajectory (A or B) actually produced
  the code patch

A valid selection instance must satisfy ALL THREE of the following
criteria:

(i) Trajectory coherence.
    Each trajectory must be internally consistent: the agent's
    reasoning, file exploration, and observed code should plausibly
    belong to a single coherent debugging rollout. There should be no
    truncation artifacts, broken tool calls, contradictory observations,
    or content that leaks the masked edit (e.g., explicit mention of
    the final patch text, or post-edit verification output that reveals
    the fix).

(ii) Validity of the ground-truth label.
     Given the code patch, exactly one of the two trajectories must be
     the unambiguous source. The bug diagnosed in the labeled
     trajectory must clearly match what the patch fixes, and the other
     trajectory must diagnose a genuinely different bug or different
     root cause that the given patch would NOT fix. If both
     trajectories' diagnoses are consistent with the patch, or if
     neither clearly is, this criterion fails.

(iii) Absence of superficial shortcut cues.
      The two trajectories should be comparable along
      content-independent dimensions, so that selection requires
      understanding the bug rather than exploiting surface statistics.
      Examples of failures: large length disparity, different
      formatting or tool-call styles, only one trajectory inspecting
      the file/function actually modified by the patch, distinctive
      tokens or identifiers from the patch appearing in only one
      trajectory, or masking applied inconsistently between the two.

Analyze the instance carefully against each criterion, then provide
your judgment in the following format:

<overall>VALID or INVALID or UNCERTAIN</overall>
<confidence>A number from 1 to 5</confidence>


Where:
- VALID: All three criteria clearly pass.
- INVALID: At least one criterion clearly fails.
- UNCERTAIN: You cannot confidently decide on at least one criterion.
- Confidence: 1 = very uncertain, 5 = very confident in your overall judgment.

Be thorough and precise in your comparison.
\end{Verbatim}
\end{promptbox}

\subsection{Contrast Image Pairs Verification Prompt}
\label{app:prompt_edit_verifier}
This is the prompt used by GPT~5.4 to verify the quality of contrast image pairs. The verifier jointly assesses visual coherence, answer validity, and preservation of question-irrelevant content, and emits an overall \texttt{VALID}/\texttt{INVALID}/\texttt{UNCERTAIN} decision together with a confidence score. Examples marked \texttt{UNCERTAIN} are escalated to manual review.
\begin{promptbox}{Contrast Image Pair Verification Prompt Template}
\begin{Verbatim}[fontsize=\small,breaklines=true,breakanywhere=true]
You are a quality verifier for a VQA dataset. You
will be shown a contrast pair created by editing an original image, and
your job is to decide whether the edited example is valid.

You will receive:
- Image 1 (Original Image, I): the source image before editing
- Image 2 (Edited Image, I'): the image after a localized edit
- Question (Q): a question about the image
- Original Answer (A): the correct answer to Q on the original image I
- New Answer (A'): the intended correct answer to Q on the edited image I'

A valid edited example must satisfy ALL THREE of the following criteria:

(i) Visual coherence of the edited image.
    The edited image I' must look natural and internally consistent.
    There should be no obvious editing artifacts, such as blurred or
    warped regions, broken object boundaries, implausible lighting or
    shadows, mismatched textures, floating objects, or physically or
    semantically incoherent content.

(ii) Validity of the new answer A' given I' and Q.
     When the question Q is asked about the edited image I', the answer
     A' must be clearly and unambiguously correct based solely on the
     visual content of I'. If A' is not supported, only partially
     supported, or if multiple answers (including the original A) remain
     plausible for I', this criterion fails.

(iii) Preservation of question-irrelevant content.
      All regions of the image that are not relevant to answering Q
      should remain essentially unchanged between I and I'. The edit
      should be localized to the answer-relevant region; unrelated
      objects, background, layout, style, and overall scene composition
      must be preserved. Global restyling, scene-wide changes, or
      unintended modifications to irrelevant regions cause this
      criterion to fail.

Analyze the pair carefully against each criterion, then provide your
judgment in the following format:

<overall>VALID or INVALID or UNCERTAIN</overall>
<confidence>A number from 1 to 5</confidence>

Where:
- VALID: All three criteria clearly pass.
- INVALID: At least one criterion clearly fails.
- UNCERTAIN: You cannot confidently decide on at least one criterion
- Confidence: 1 = very uncertain, 5 = very confident in your overall
  judgment.

Be thorough and precise in your comparison.
\end{Verbatim}
\end{promptbox}

\subsection{Agentic Coding Prompt}
\label{app:prompt_agentic_coding}

This is the prompt used for the agentic coding setting. It consists of a system message that fixes the response format and a per-instance template that wraps the PR description and task instructions. \texttt{\{task\}} is filled with the natural-language PR description for the SWE-Bench instance.

\begin{promptbox}{Agentic Coding System Prompt}
\begin{Verbatim}[breaklines=true,breakanywhere=true,fontsize=\small]
You are a helpful assistant that can interact multiple times with a computer
shell to solve programming tasks.
Your response must contain exactly ONE bash code block with ONE command (or
commands connected with && or ||).

Include a THOUGHT section before your command where you explain your reasoning
process.
Format your response as shown in <format_example>.

<format_example>
THOUGHT: Your reasoning and analysis here

```bash
your_command_here
```
</format_example>

Failure to follow these rules will cause your response to be rejected.
\end{Verbatim}
\end{promptbox}

\begin{promptbox}{Agentic Coding Instance Prompt Template}
\begin{Verbatim}[breaklines=true,breakanywhere=true,fontsize=\small]
<pr_description>
Consider the following PR description:
{task}
</pr_description>

<instructions>
# Task Instructions

## Overview
You're a software engineer interacting continuously with a computer by submitting
commands.
You'll be helping implement necessary changes to meet requirements in the PR
description.
Your task is specifically to make changes to non-test files in the current
directory in order to fix the issue described in the PR description in a way
that is general and consistent with the codebase.

IMPORTANT: This is an interactive process where you will think and issue ONE
command, see its result, then think and issue your next command.

For each response:
1. Include a THOUGHT section explaining your reasoning and what you're trying
   to accomplish
2. Provide exactly ONE bash command to execute

## Important Boundaries
- MODIFY: Regular source code files in /testbed (this is the working directory
  for all your subsequent commands)
- DO NOT MODIFY: Tests, configuration files (pyproject.toml, setup.cfg, etc.)

## Recommended Workflow
1. Analyze the codebase by finding and reading relevant files
2. Create a script to reproduce the issue
3. Edit the source code to resolve the issue
4. Verify your fix works by running your script again
5. Test edge cases to ensure your fix is robust

## Command Execution Rules
You are operating in an environment where
1. You write a single command
2. The system executes that command in a subshell
3. You see the result
4. You write your next command

Each response should include:
1. A **THOUGHT** section where you explain your reasoning and plan
2. A single bash code block with your command

Format your responses like this:

<format_example>
THOUGHT: Here I explain my reasoning process, analysis of the current situation,
and what I'm trying to accomplish with the command below.

```bash
your_command_here
```
</format_example>

Commands must be specified in a single bash code block:

```bash
your_command_here
```

**CRITICAL REQUIREMENTS:**
- Your response SHOULD include a THOUGHT section explaining your reasoning
- Your response MUST include EXACTLY ONE bash code block
- This bash block MUST contain EXACTLY ONE command (or a set of commands
  connected with && or ||)
- If you include zero or multiple bash blocks, or no command at all, YOUR
  RESPONSE WILL FAIL
- Do NOT try to run multiple independent commands in separate blocks in one
  response
- Directory or environment variable changes are not persistent. Every action is
  executed in a new subshell.
- However, you can prefix any action with
  `MY_ENV_VAR=MY_VALUE cd /path/to/working/dir && ...` or write/load
  environment variables from files

## Environment Details
- You have a full Linux shell environment
- Always use non-interactive flags (-y, -f) for commands
- Avoid interactive tools like vi, nano, or any that require user input
- If a command isn't available, you can install it

## Useful Command Examples

### Create a new file:
```bash
cat <<'EOF' > newfile.py
import numpy as np
hello = "world"
print(hello)
EOF
```

### Edit files with sed:
```bash
# Replace all occurrences
sed -i 's/old_string/new_string/g' filename.py
```

### View file content:
```bash
# View specific lines with numbers
nl -ba filename.py | sed -n '10,20p'
```

## Submission
When you've completed your work (reading, editing, testing), and cannot make
further progress issue exactly the following command:

```bash
echo MINI_SWE_AGENT_FINAL_OUTPUT && git add -A && git diff --cached
```

This command will submit your work.
You cannot continue working (reading, editing, testing) in any way on this task
after submitting.
</instructions>
\end{Verbatim}
\end{promptbox}

\subsection{Multimodal Reasoning Prompt}
\label{app:prompt_mm_reasoning}

This is the format prompt used during GRPO training and evaluation in the multimodal reasoning setting. The model is required to produce its chain of thought and final answer in dedicated tags.

\begin{promptbox}{Multimodal Reasoning Prompt Template}
\begin{Verbatim}[fontsize=\small]
{Question}
{Options}
To answer this question, you need to think step by step. Provide your thinking
process in <think></think>. Provide your final answer in <answer></answer>.
\end{Verbatim}
\end{promptbox}

%=====================================================================
\section{Training Details and Hyperparameters}
\label{app:training_details}
%=====================================================================

\subsection{Agentic Setting}
\label{app:hp_agentic}

We train with GRPO implemented in the \textbf{SkyRL} framework~\citep{serrano2023skrl}, which is designed for long-horizon tool-use rollouts in sandboxed coding environments. We adopt mini-SWE-agent~\citep{yang2024sweagent} as the agent scaffold for both training rollouts and evaluation. The training set comprises 8k instances in total: 7k standard agentic coding tasks drawn from SWE-Gym~\citep{pan2025trainingsoftwareengineeringagents} for $\mathcal{L}_{\text{GRPO}}$, and 1k contrastive trajectory pairs constructed by the pipeline in \S\ref{sec:data_agentic} for $\mathcal{L}_{\text{CA}}$. To match the total amount of training data, the RL baseline is trained on 8k SWE-Gym tasks (the same 7k plus 1k additional samples drawn from SWE-Gym in lieu of contrastive pairs) for the same number of steps. SWE-Gym instances are disjoint from SWE-Bench Verified and SWE-Bench Lite by construction, and we manually verified that no training instance overlaps with either evaluation set. The context-selection data uses the prompt in \S\ref{app:prompt_traj_choose}; for Klear-AgentForge-8B, we prefill the response with \texttt{"The correct option is"} so that the next-token logit aligned to the answer letter is well-defined, while for Qwen3-8B no prefill is needed. Table~\ref{tab:hp_agentic} lists the key hyperparameters.

\begin{table}[htbp]
\centering
\small
\setlength{\tabcolsep}{8pt}
\begin{tabular}{l l}
\toprule
\textbf{Hyperparameter} & \textbf{Value} \\
\midrule
KL loss / coefficient       & enabled, $\beta = 1.0\times 10^{-3}$, $k_3$ estimator \\
Optimizer / learning rate   & AdamW, $2.0\times 10^{-6}$ \\
Train batch size            & $16$ \\
Policy mini-batch           & $8$ \\
Rollouts per prompt $G$     & $8$ \\
Max prompt length per turn  & $4{,}096$  \\
Max response length per turn & $4{,}096$ \\
Max input length (full ctx) & $28{,}672$  \\
Max agent turns             & $35$ \\
CA batch size               & $8$  \\
CA max prompt length        & $32{,}768$ \\
CA margin clip $c$          & $5.0$ \\
CA coefficient $\lambda$    & $0.005$ (Klear-AgentForge-8B) \,/\, $0.001$ (Qwen3-8B) \\
Total epochs                & $3$ \\
\bottomrule
\end{tabular}
\caption{Key training hyperparameters for the agentic coding setting.}
\label{tab:hp_agentic}
\end{table}

\subsection{Multimodal Setting}
\label{app:hp_multimodal}

We train with the GRPO algorithm implemented in the Easy-R1 framework~\citep{zheng2025easyr1}. The training set consists of 45k examples: 38k standard single-image QA examples for $\mathcal{L}_{\text{GRPO}}$ and 7k contrastive image pairs for $\mathcal{L}_{\text{CA}}$. To match the total amount of training data, the RL baseline is trained on 45k standard QA examples (the same 38k plus 7k additional samples drawn from the same source pool in lieu of contrastive pairs) for the same number of steps. For source datasets that also appear in our evaluation suite (notably ScienceQA), training draws exclusively from the official train split while evaluation uses the official test split; we manually verified that no training example overlaps with any evaluation benchmark. Both Qwen2.5-VL-7B and Qwen3-VL-8B share the same backbone configuration; the only base-model-specific difference is the $\mathcal{L}_{\mathrm{CA}}$ coefficient $\lambda$, selected from the ablation in \S\ref{sec:abl_lambda}. For Qwen3-VL-8B we additionally override the chat template with the official thinking template so that rollouts follow the \texttt{<think>...</think>} format, and use a slightly relaxed validation sampler (temperature $0.6$, top-$p$ $0.95$). The context-selection data uses the prompt in \S\ref{app:prompt_image_choose}. For both Qwen2.5-VL-7B and Qwen3-VL-8B, no response prefill is needed. Table~\ref{tab:hp_multimodal} lists the key hyperparameters.

\begin{table}[ht]
\centering
\small
\setlength{\tabcolsep}{8pt}
\begin{tabular}{l l}
\toprule
\textbf{Hyperparameter} & \textbf{Value} \\
\midrule
KL coefficient $\beta$     & $1.0\times 10^{-2}$ \\
Optimizer                  & AdamW, weight decay $1.0\times 10^{-2}$ \\
Learning rate              & $1.0\times 10^{-6}$ \\
Max gradient norm          & $1.0$ \\
Rollout batch size         & $256$ \\
Actor mini-batch size      & $64$ \\
Rollouts per prompt $G$    & $8$ \\
Rollout temperature        & $1.0$ \\
Max prompt length          & $16{,}384$ \\
Max response length        & $4{,}096$  \\
CA batch size              & $32$\\
CA margin clip $c$         & $5.0$ \\
CA coefficient $\lambda$   & $0.005$ (Qwen2.5-VL-7B) \,/\, $0.001$ (Qwen3-VL-8B) \\
Total epochs               & $3$ \\
\bottomrule
\end{tabular}
\caption{Key training hyperparameters for the multimodal setting.}
\label{tab:hp_multimodal}
\end{table}

%=====================================================================
\section{Compute Resources}
\label{app:compute}
%=====================================================================

\paragraph{Hardware.}
All RL training runs in this paper are conducted on a single node equipped with $4\times$ NVIDIA H200 (140GB HBM3e) GPUs, intra-node NVLink interconnect, and $\geq 500$ GB host RAM. The same hardware configuration is used for both the multimodal and agentic settings.

\paragraph{Per-experiment GPU time.}
Table~\ref{tab:compute_per_run} reports the approximate wall-clock and GPU-hour cost per training run on the hardware above. ``Per-experiment'' refers to a single end-to-end RL training run for one base model under one configuration (\eg, one entry in Table~\ref{tab:main_results} or Table~\ref{tab:agentic_main}); ablation studies (\S\ref{sec:ablations}, \S\ref{sec:abl_lambda_agentic}) and the data-augmentation comparisons (\S\ref{sec:vs_data_aug}) each consume an additional run of comparable cost.

\begin{table}[ht]
\centering
\small
\setlength{\tabcolsep}{8pt}
\begin{tabular}{l l c c}
\toprule
\textbf{Setting} & \textbf{Base model} & \textbf{Wall-clock (h)} & \textbf{GPU-hours} \\
\midrule
\multirow{2}{*}{Multimodal} & Qwen2.5-VL-7B & $\sim 60$  & $\sim 240$ \\
                            & Qwen3-VL-8B   & $\sim 72$  & $\sim 288$ \\
\midrule
\multirow{2}{*}{Agentic}    & Klear-AgentForge-8B-SFT & $\sim 72$ & $\sim 288$ \\
                            & Qwen3-8B                & $\sim 72$ & $\sim 288$ \\
\bottomrule
\end{tabular}
\caption{Approximate per-run compute cost on $4\times$ H200. GPU-hours are wall-clock $\times 4$.}
\label{tab:compute_per_run}
\end{table}

\paragraph{Foundation-model inference cost during data construction.}
Nano Banana 2 was queried approximately 10k times for generative editing of natural images; GPT-5.4 was queried approximately 10k times as the automatic verifier of edited images.

\section{Ablations}
\label{sec:ablations}
In this section we provide detailed ablation studies for both agentic and multimodal settings.

\subsection{Agentic Setting}
\label{app:ablation_agentic}

\paragraph{Effect of context-awareness loss coefficient $\lambda$.}
\label{sec:abl_lambda_agentic}
\begin{wraptable}{r}{0.35\linewidth}
\centering\small
\setlength{\tabcolsep}{4pt}
\begin{tabular}{lcc}
\toprule
$\lambda$ & Verified & Lite \\
\midrule
0.001 & 28.2 & 21.3 \\
 0.005 & \textbf{30.2} & \textbf{24.0} \\
0.01  & 28.2 & 20.0 \\
\bottomrule
\end{tabular}
\caption{\small Effect of $\lambda$ on agentic coding. Resolve rate (\%) on SWE-Bench.}
\label{tab:agentic_lambda}
\end{wraptable}
We study the sensitivity of the agentic setting to the weight $\lambda$ of the context-awareness loss in Eq.~\ref{eq:combined}, sweeping $\lambda \in \{0.001, 0.005, 0.01\}$ on Klear-AgentForge-8B. Table~\ref{tab:agentic_lambda} reveals a clear trend: performance peaks at $\lambda=0.005$. At $\lambda=0.001$, the resolve rate is indistinguishable from the RL baseline reported in Table~\ref{tab:agentic_main}, indicating that the context-awareness signal is too weak to meaningfully influence training. At $\lambda=0.01$, the score drops below even the RL baseline, suggesting that an overly strong CA term competes with the GRPO objective and degrades final task performance. We adopt $\lambda=0.005$ as the default for agentic training.

\subsection{Multimodal Setting}
\label{app:ablation_multimodal}

We conduct ablation studies to understand the sensitivity of our method to three key hyperparameters: the proportion of contrastive data, the maximum response length during RL rollouts, and the contrastive loss coefficient $\lambda$.

\begin{figure}[ht]
\centering
\includegraphics[width=\linewidth]{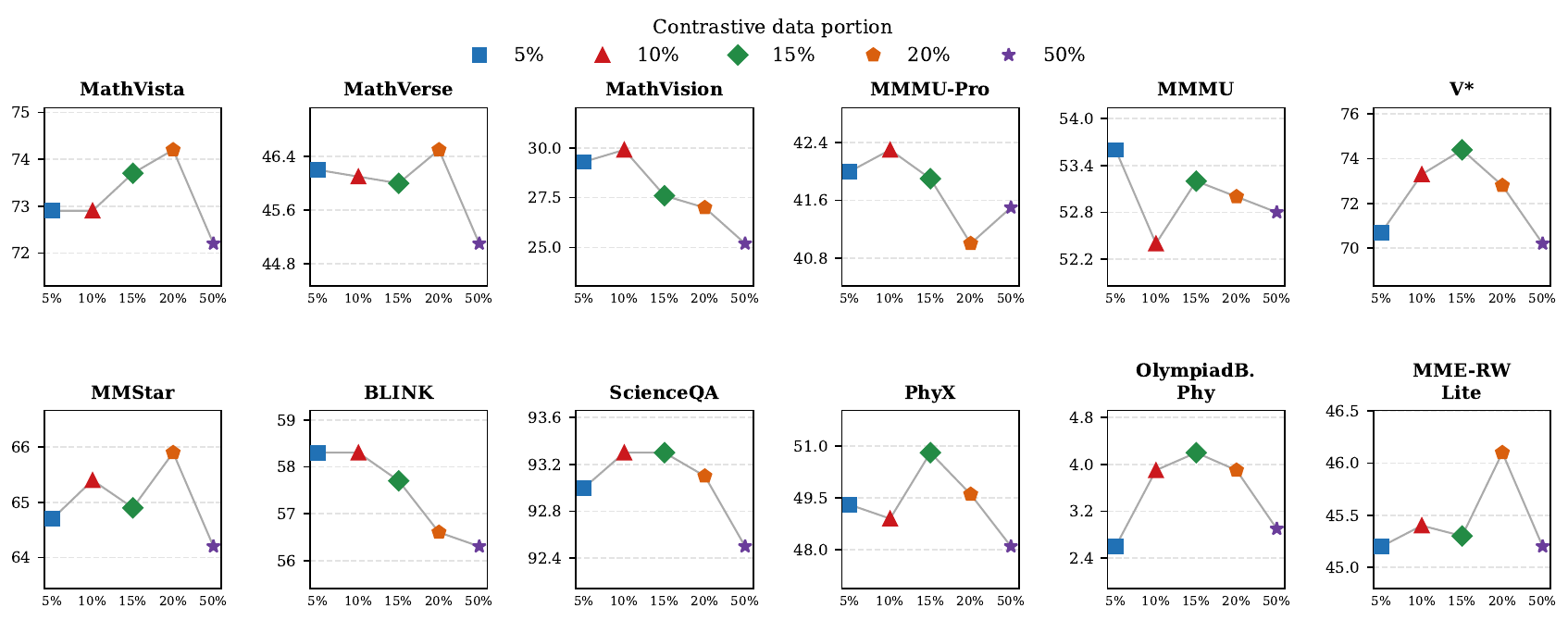}
\caption{Effect of contrastive data proportion on Qwen2.5-VL-7B performance. The x-axis is data proportion and the y-axis is accuracy (\%). The 15\% setting achieves the best overall balance.}
\label{fig:ablation_portion}
\end{figure}

\paragraph{Effect of contrastive data proportion.}
\label{sec:abl_portion}
We vary the proportion of contrastive image pairs in the training set across five levels: 5\%, 10\%, 15\%, 20\%, and 50\%. Using Qwen2.5-VL-7B as the base model, we maintain the same total training steps across all five settings. Figure~\ref{fig:ablation_portion} visualizes performance across all 12 benchmarks.

The results reveal a clear inverted-U pattern. Increasing the contrastive proportion from 5\% to 15\% generally improves performance on most benchmarks. However, further increasing to 20\% and especially 50\% degrades performance broadly: at 50\%, nearly all benchmarks are at or below the 5\% level. This trend indicates that excessive contrastive data displaces the standard reasoning supervision. We select 15\% as our default setting, as it strikes the best balance between context sensitivity and reasoning capability.

\begin{figure}[ht]
\centering
\includegraphics[width=\linewidth]{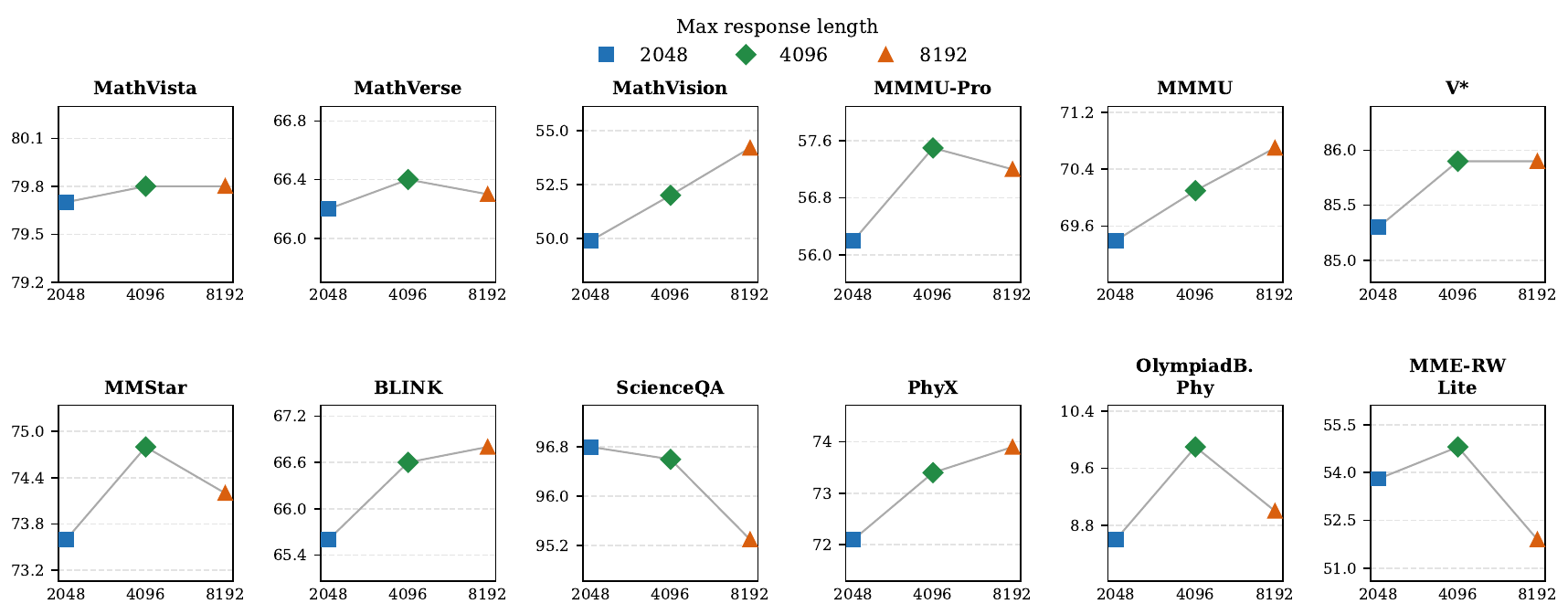}
\caption{Effect of maximum response length on Qwen3-VL-8B performance. Max length 4096 offers the best average performance across all benchmarks.}
\label{fig:ablation_length}
\end{figure}

\paragraph{Effect of maximum response length.}
\label{sec:abl_length}
During RL rollouts, the maximum response length controls the reasoning budget available to the model. We compare three settings, 2048, 4096, and 8192 tokens, on Qwen3-VL-8B. Figure~\ref{fig:ablation_length} shows the results across all 12 benchmarks.

From figure~\ref{fig:ablation_length} we can find that increasing the response length from 2048 to 4096 yields consistent improvements across most benchmarks, with notable gains on tasks requiring extended reasoning chains like MathVision and MMMU-Pro. Further increasing to 8192 provides additional gains on MathVision and MMMU, but slightly hurts performance on ScienceQA, OlympiadBench Physics, and MME-RealWorld, likely because longer generation windows introduce noise on tasks that do not benefit from extended reasoning. We select 4096 as our default, as it offers the best average performance across all benchmarks.

\begin{figure}[ht]
\centering
\includegraphics[width=\linewidth]{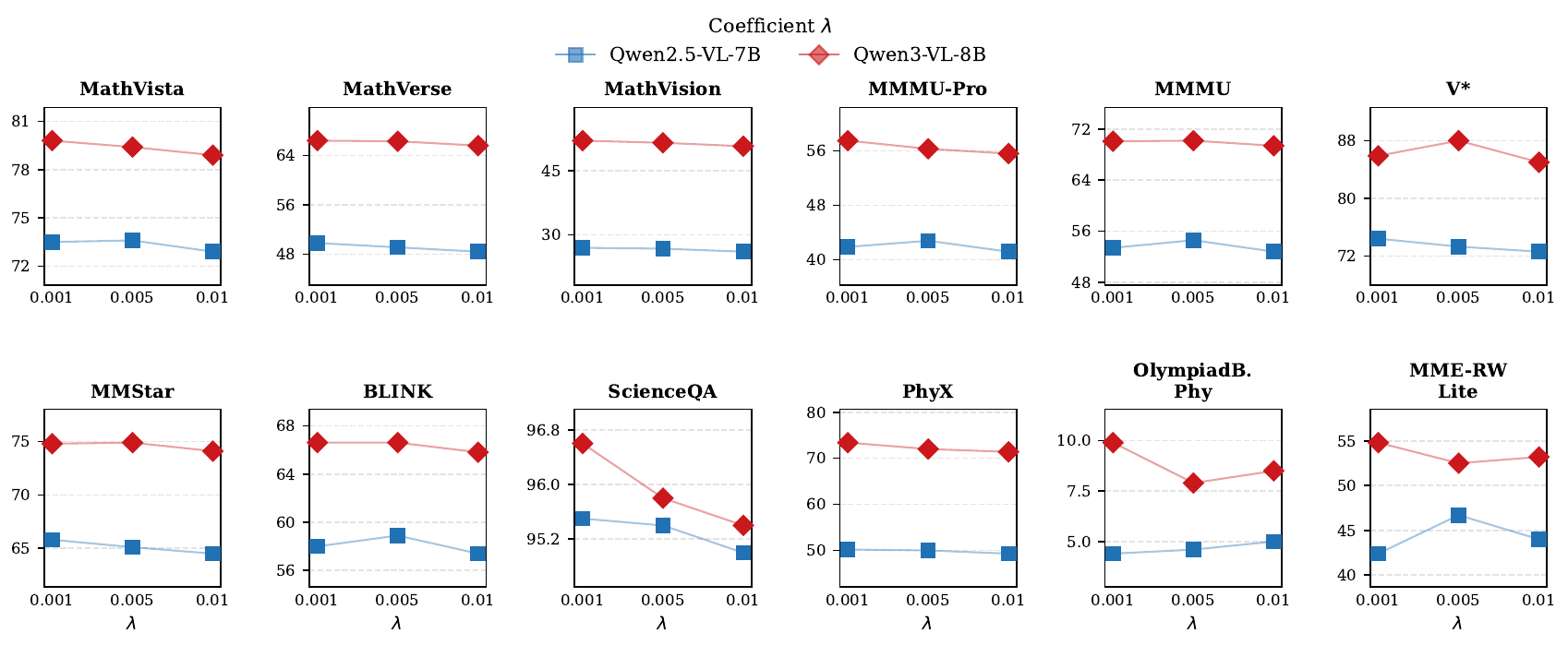}
\caption{Effect of the contrastive loss coefficient $\lambda$ on benchmark performance. Both models are stable across small $\lambda$ values but degrade on the majority of benchmarks once $\lambda$ is increased to 0.01.}
\label{fig:ablation_lambda}
\end{figure}

\paragraph{Effect of contrastive loss coefficient $\lambda$.}
\label{sec:abl_lambda}
The coefficient $\lambda$ in Eq.~\ref{eq:combined} controls the relative weight of the context-selection loss. We compare $\lambda \in \{0.001, 0.005, 0.01\}$ on both Qwen2.5-VL-7B and Qwen3-VL-8B. Figure~\ref{fig:ablation_lambda} shows per-benchmark results.

Within the small-$\lambda$ regime, both models are relatively robust to the choice between $0.001$ and $0.005$. On Qwen2.5-VL-7B, $\lambda = 0.005$ achieves slightly better overall performance, while on Qwen3-VL-8B, $\lambda = 0.001$ performs slightly better on average. Further increasing $\lambda$ to $0.01$, however, degrades performance on the majority of benchmarks for both models. This matches our expectation: the context-selection loss is intended as an auxiliary objective that augments rather than competes with the RL objective. When $\lambda$ becomes too large, the CA gradient overwhelms the GRPO update signal, and the auxiliary signal ends up doing more harm than good. We therefore select $\lambda = 0.005$ for Qwen2.5-VL-7B and $\lambda = 0.001$ for Qwen3-VL-8B.

\section{Limitations}
\label{sec:limitation}
Due to computational constraints, all of our experiments are conducted on base models with fewer than 10B parameters; we have not validated \ours\ at substantially larger scales (e.g., 30B+ or 70B+).  Moreover, the majority of base models we evaluate are drawn from the Qwen family. Validating \ours\ across a broader set of model families would strengthen the generality of our claims.

%=====================================================================
\section{Broader Impacts}
\label{app:broader_impacts}
%=====================================================================
We propose a new RL post-training paradigm that explicitly rewards context grounding rather than only the final answer, with the aim of making both multimodal and agentic LLMs more faithful to the evidence in their inputs. We expect the positive impact to come from improved reliability in tasks where shortcut answers are dangerous: software-engineering agents that must read code carefully before editing it, and visual reasoning systems that must read fine-grained perceptual evidence before answering. Like any technique that strengthens general LLM capabilities, \ours\ could in principle be used to amplify harmful applications, but it does not require any additional sensitive data. We do not anticipate negative societal impacts.

\end{document}